%% file: main.tex
\documentclass[sigconf]{acmart}
\setcopyright{none}                                   
\renewcommand\footnotetextcopyrightpermission[1]{}   
\usepackage{xcolor}
\usepackage{soul}
\usepackage{booktabs}
\usepackage{multirow}
\usepackage{subcaption}
\usepackage{makecell}
\usepackage{adjustbox}
\usepackage{subcaption}
\usepackage{xspace}
\newcommand{\sys}{\textsc{XposedMIA}\xspace}
\newcommand{\descr}[1]{\smallskip \noindent \textbf{#1}}
\newcommand{\descrit}[1]{\smallskip \noindent}
\usepackage{array}
\usepackage{threeparttable}
\usepackage{tabularx}
\usepackage{cleveref}
\usepackage{capt-of}
\usepackage{placeins}
\usepackage{float}
\usepackage{soul}
\usepackage[ruled,vlined,linesnumbered]{algorithm2e}

\usepackage{amsmath,amssymb}

\usepackage{tikz}
\usepackage{pgfplots}
\pgfplotsset{compat=1.18}
\usepackage{xcolor}

\AtBeginDocument{%
  }

\setcopyright{acmlicensed}
\copyrightyear{2018}
\acmYear{2018}
\acmDOI{XXXXXXX.XXXXXXX}
\acmConference[Conference acronym 'XX]{Make sure to enter the correct
  conference title from your rights confirmation email}{June 03--05,
  2018}{Woodstock, NY}
\acmISBN{978-1-4503-XXXX-X/2018/06}

\begin{document}

\title{XposedMIA: Exploiting Explainable Artificial Intelligence (XAI) for One-Shot Membership Inference Attacks}

\author{Omar Hamdache}
\email{omar.hamdache@bilkent.edu.tr}
\affiliation{%
  \institution{Bilkent University}
  \city{Ankara}
  \country{Turkey}
}

\author{Serdar Kara}
\authornote{Both authors contributed equally to this research.}
\email{serdar.kara@ug.bilkent.edu.tr}
\affiliation{%
  \institution{Bilkent University}
  \city{Ankara}
  \country{Turkey}
}

\author{Kıvanç Kuzey Dikici}
\authornotemark[1]
\email{kuzey.dikici@ug.bilkent.edu.tr}
\affiliation{%
  \institution{Bilkent University}
  \city{Ankara}
  \country{Turkey}
}

\author{Abdullah Caglar Oksuz}
\email{axo223@case.edu}
\affiliation{%
  \institution{Case Western Reserve University}
  \city{Cleveland}
  \state{Ohio}
  \country{USA}
}

\author{Anisa Halimi}
\email{anisa.halimi@ibm.com}
\affiliation{%
  \institution{IBM Research}
  \city{Dublin}
  \country{Ireland}
}

\author{Erman Ayday}
\email{exa208@case.edu}
\affiliation{%
  \institution{Case Western Reserve University}
  \city{Cleveland}
  \state{Ohio}
  \country{USA}
}

\author{Sinem Sav}
\authornote{Corresponding author.}
\email{sinem.sav@cs.bilkent.edu.tr}
\affiliation{%
  \institution{Bilkent University}
  \city{Ankara}
  \country{Turkey}
}

\renewcommand{\shortauthors}{Hamdache et al.}

\begin{abstract}
  Machine learning (ML) services increasingly expose explainability outputs alongside model predictions. While such explanations are intended to improve transparency, they may also reveal information about the underlying training data. In this work, we show that the local surrogate geometry employed by Local Interpretable Model-agnostic Explanations (LIME) can be repurposed as a proxy for a sample's distance to the model's decision boundary. We extend this boundary-distance view to tabular counterfactual explanations and, for the first time, to visual counterfactual explanations for image classifiers. We present \sys, a one-shot, explanation-based black-box membership inference attack (MIA) framework that converts the returned explanation into a proxy for the queried sample's distance to the model decision boundary. After a one-time calibration step on a small labeled calibration set of as few as $\sim30$ samples, our attack requires only a single query for each target sample. Unlike prior label-only attacks requiring iterative queries or shadow training, our method relies solely on the predicted label and its accompanying explanation. Our evaluation indicates that \sys achieves high effectiveness, with success rates of up to $0.95$ using only a single query per target, revealing that releasing explanations introduces an additional channel for membership leakage.
\end{abstract}





\maketitle

\input{sections/1-introduction}
\input{sections/2-related_work}

\input{sections/3-background}
\input{sections/4-threat}
\input{sections/5-method}
\input{sections/6-results}
\input{sections/7-dicussion}
\input{sections/8-conclusion}

\begin{acks}
This work was supported in part by the National Science Foundation (NSF) under Grants OAC-2112606, 2427505, and 2141622; the National Institutes of Health (NIH) under Grant R01LM014520; and the U.S. Department of Agriculture (USDA) under Grant 2026-67023-46766. 
\end{acks}



\bibliographystyle{ACM-Reference-Format}
\bibliography{bibfile}

\appendix

\section{Distance Threshold Calibration (DTC)}
\label{app:dtc}
Algorithm~\ref{alg:calibration} formalizes the two-stage calibration procedure described in Section~\ref{sec:calibration}.

\renewcommand{\thealgocf}{\thesection.\arabic{algocf}}
\setcounter{algocf}{0}

\begin{algorithm}[h]
    \caption{Distance Threshold Calibration (DTC)}
    \label{alg:calibration}

    \KwIn{Calibration records
    $\mathcal{B}_q=
    \{(d_i^{(\mathcal{E},q)},m_i)\}$ for distance measure $q$,
    containing finite boundary-distance estimates
    $d_i^{(\mathcal{E},q)}$ and known membership labels
    $m_i\in\{0,1\}$}

    \KwOut{Calibrated threshold $\tau_q$ and inequality direction
    $\mathit{dir}_q$}

    \BlankLine
    $\mathcal{T}_q
    \leftarrow
    \operatorname{sort}
    \left(
    \operatorname{unique}
    \left(
    \{d\mid(d,m)\in\mathcal{B}_q\}
    \right)
    \right)$\;

    $N_{\mathrm{mem}}
    \leftarrow
    |\{(d,m)\in\mathcal{B}_q\mid m=1\}|$\;

    $N_{\mathrm{non}}
    \leftarrow
    |\{(d,m)\in\mathcal{B}_q\mid m=0\}|$\;

    \BlankLine
    $\mathit{best\_acc}\leftarrow-\infty$\;

    \ForEach{$\tau_c\in\mathcal{T}_q$}{
        \ForEach{$\mathit{dir}\in
        \{\mathtt{leq},\mathtt{gt}\}$}{

            Compute $\mathit{acc}_{\mathrm{mem}}
            (\tau_c,\mathit{dir})$ and
            $\mathit{acc}_{\mathrm{non}}
            (\tau_c,\mathit{dir})$\;

            $\mathit{acc}_{C}
            \leftarrow
            \dfrac{
            N_{\mathrm{mem}}\,
            \mathit{acc}_{\mathrm{mem}}
            +
            N_{\mathrm{non}}\,
            \mathit{acc}_{\mathrm{non}}
            }{
            N_{\mathrm{mem}}+N_{\mathrm{non}}
            }$\;

            \If{$\mathit{acc}_{C}>\mathit{best\_acc}$}{
                $\mathit{best\_acc}
                \leftarrow
                \mathit{acc}_{C}$\;

                $\mathit{dir}_q
                \leftarrow
                \mathit{dir}$\;
            }
        }
    }

    \BlankLine
    $\mathit{min\_diff}\leftarrow\infty$\;

    \ForEach{$\tau_c\in\mathcal{T}_q$}{

        Compute $\mathit{acc}_{\mathrm{mem}}
        (\tau_c,\mathit{dir}_q)$ and
        $\mathit{acc}_{\mathrm{non}}
        (\tau_c,\mathit{dir}_q)$\;

        $\mathit{diff}
        \leftarrow
        \left|
        \mathit{acc}_{\mathrm{mem}}
        -
        \mathit{acc}_{\mathrm{non}}
        \right|$\;

        \If{$\mathit{diff}<\mathit{min\_diff}$}{
            $\mathit{min\_diff}
            \leftarrow
            \mathit{diff}$\;

            $\tau_q
            \leftarrow
            \tau_c$\;
        }
    }

    \BlankLine
    \Return{$\tau_q,\mathit{dir}_q$}\;
\end{algorithm}

\section{Example Visual Counterfactual Explanations}
\label{app:visual-cf-exampe}

\begin{figure}[t]
    \centering
    \includegraphics[width=\columnwidth]{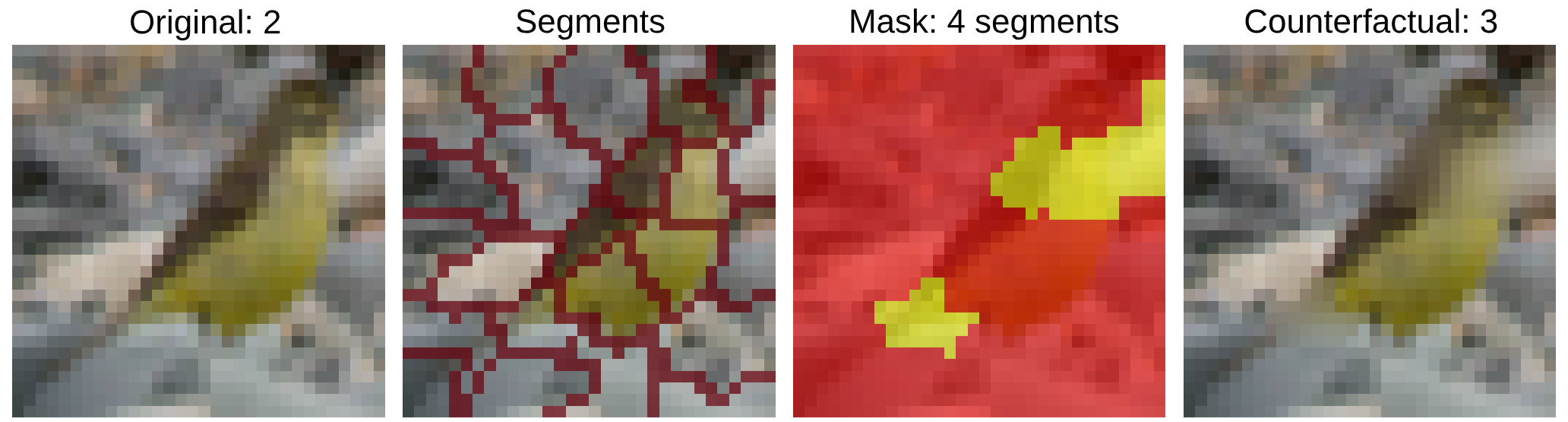}
    \caption{SEDC counterfactual example of a sample from CIFAR-10 whose class changed from the correctly classified class 2 (bird) to class 3 (cat) by applying Gaussian blur to 4 segments.}
    \label{fig:sedc-example}
\end{figure}

\begin{figure}[t]
    \centering
    \includegraphics[width=\columnwidth]{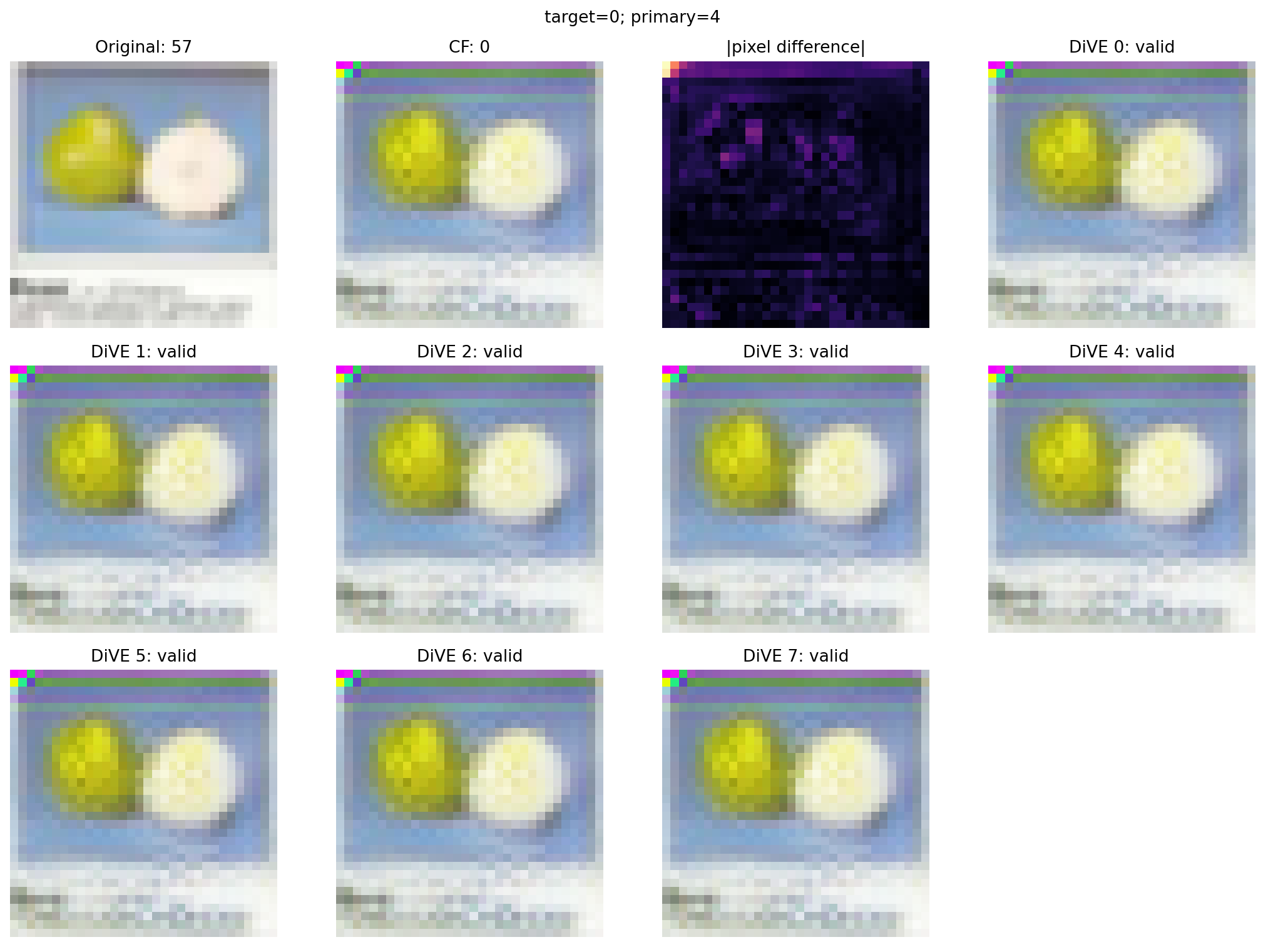}
    \caption{DiVE counterfactual example of a sample from CIFAR-100 whose class changed from the correctly classified class 57 (pear) to class 0 (apple). The primary example was chosen to be the one with index 4. DiVE generates multiple candidates and chooses the best one as previously explained in \Cref{sec:distance}.}
    \label{fig:dive-example}
\end{figure}

An example SEDC counterfactual where we used Gaussian blur to remove image segments can be seen in \Cref{fig:sedc-example}. Additionally, a DiVE example is also presented in \Cref{fig:dive-example}.

\section{Additional Ablation Studies}
\label{app:additional-ablations}

\subsection{Distance to the Decision Boundary Distribution}
\label{app:distances}

\begin{figure}[t]
    \centering

    \begin{subfigure}[b]{\columnwidth}
        \centering
        \includegraphics[width=\linewidth]{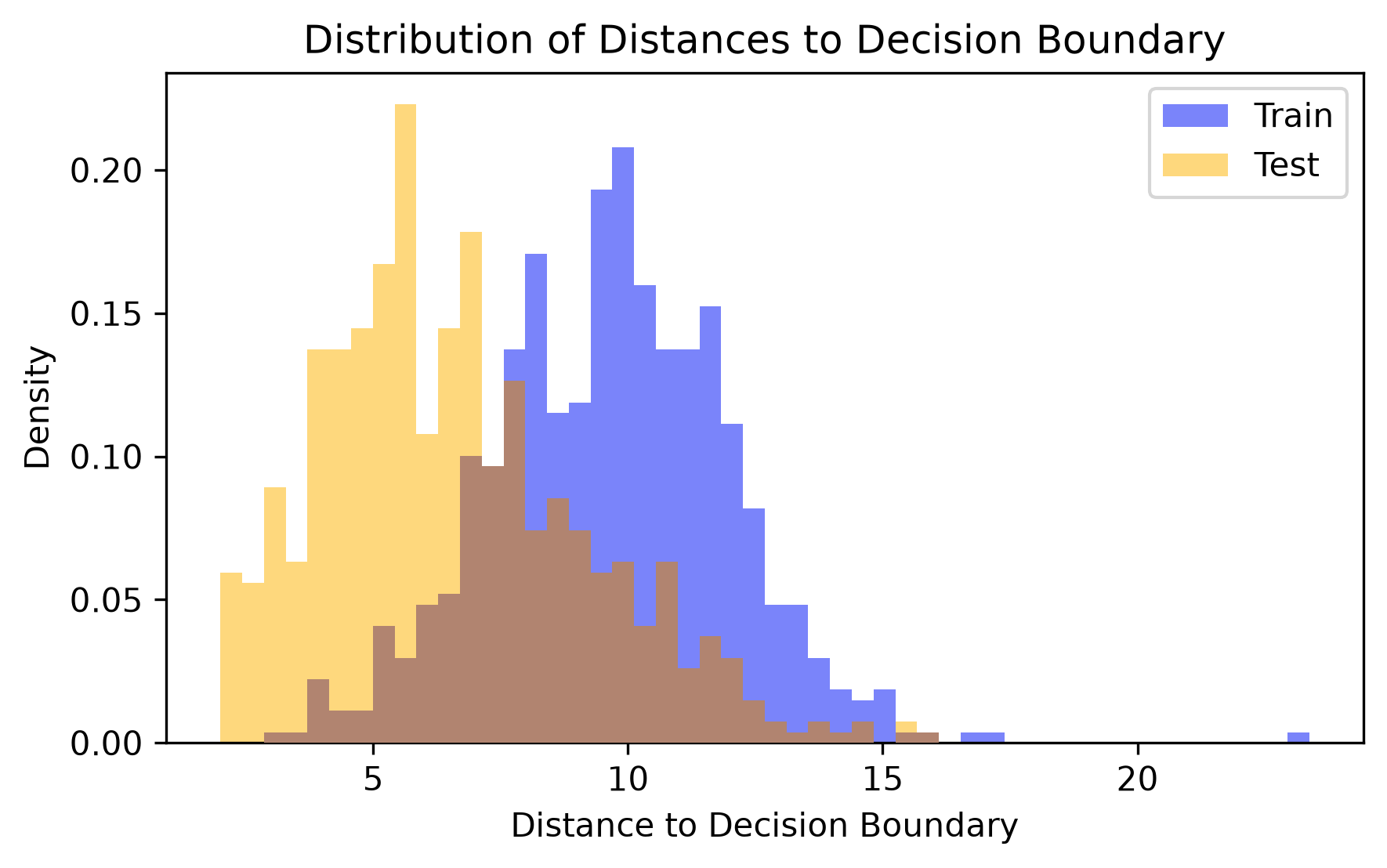}
        \caption{NICE on Location.}
        \label{fig:cf-nice-location-dist}
    \end{subfigure}

    \vspace{0.5em}

    \begin{subfigure}[b]{\columnwidth}
        \centering
        \includegraphics[width=\linewidth]{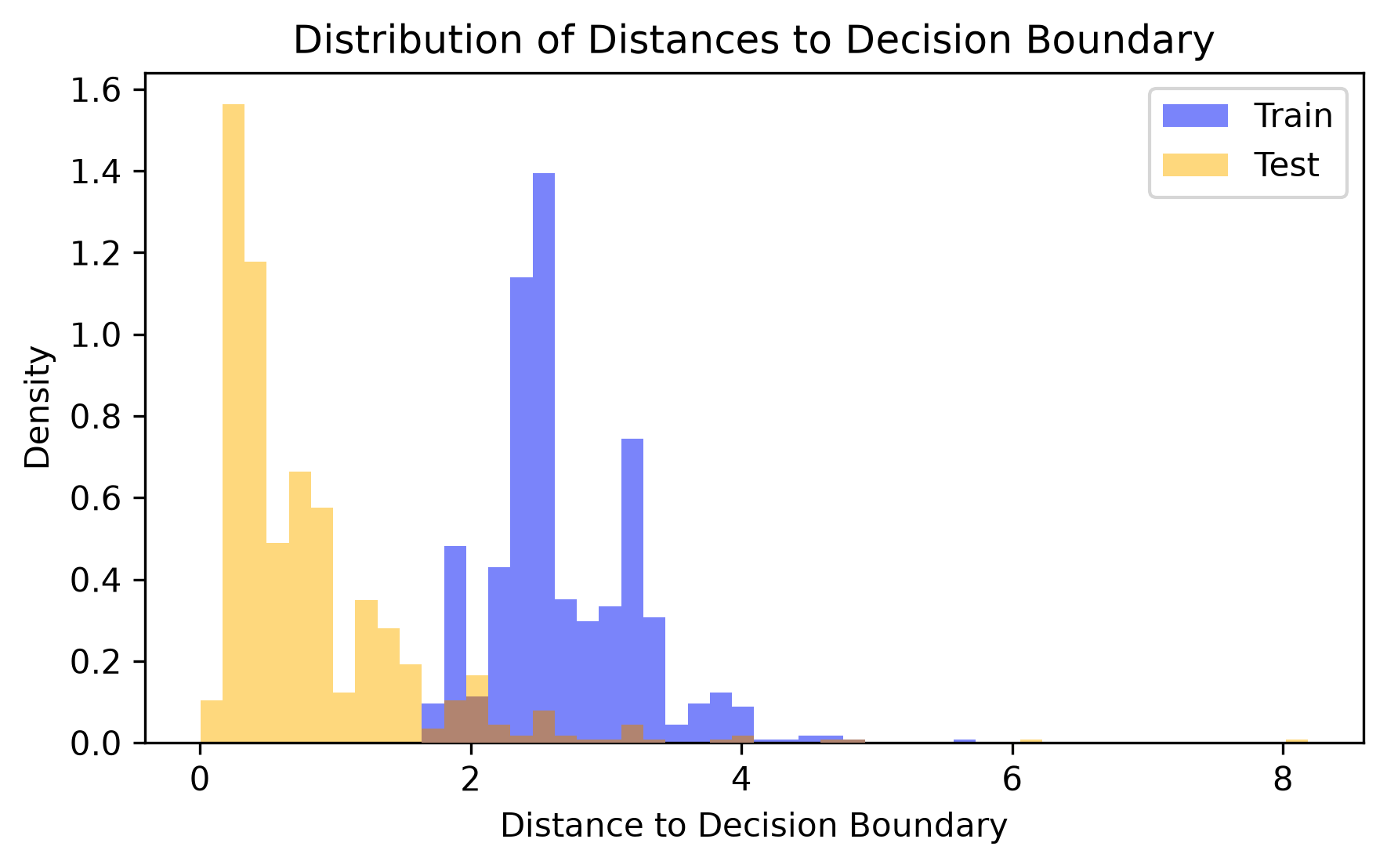}
        \caption{Wachter on Purchase-100.}
        \label{fig:cf-wachter-purchase-dist}
    \end{subfigure}

    \vspace{0.5em}

    \begin{subfigure}[b]{\columnwidth}
        \centering
        \includegraphics[width=\linewidth]{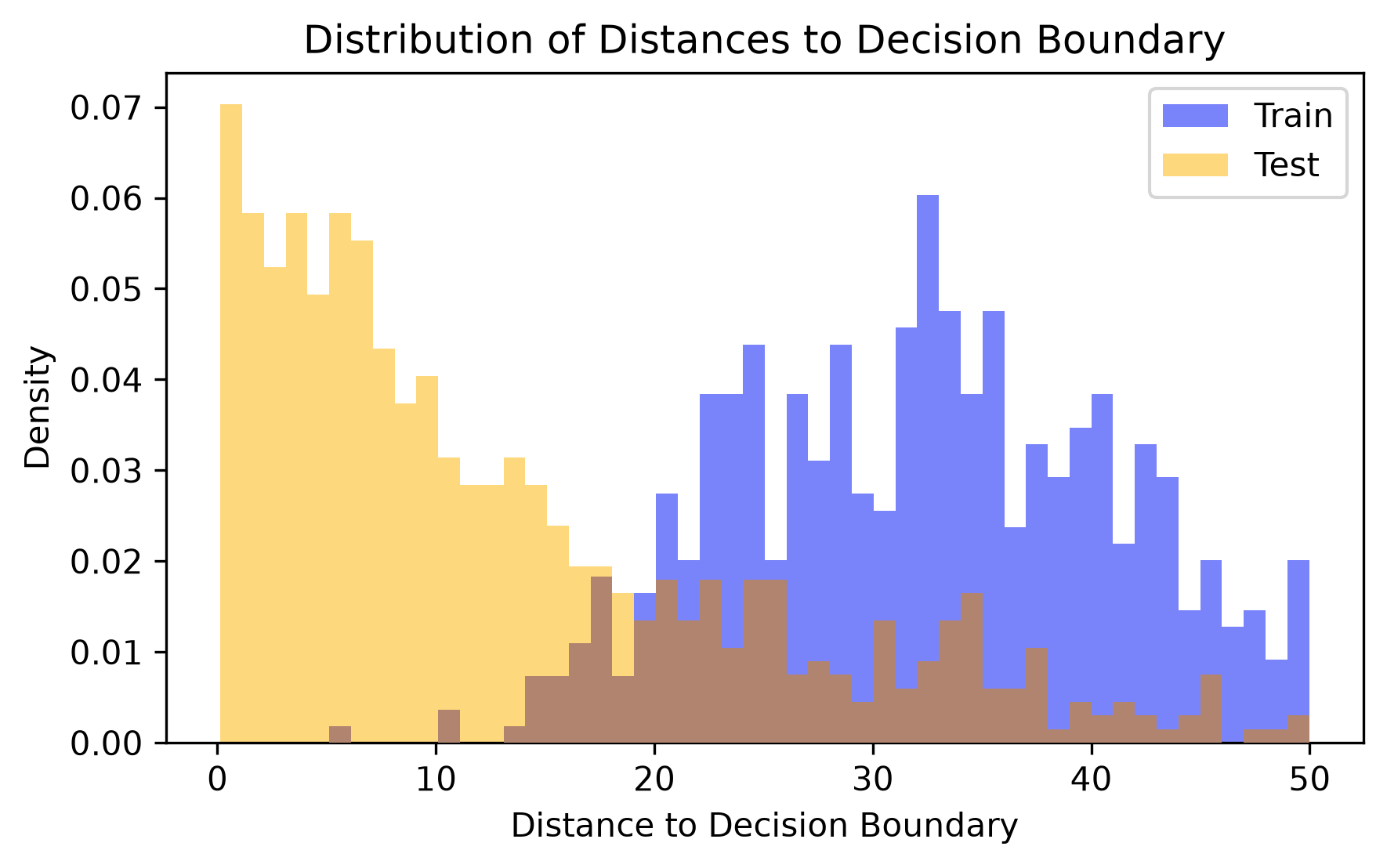}
        \caption{Growing Spheres on Texas-100.}
        \label{fig:cf-gs-texas-dist}
    \end{subfigure}

    \caption{Member and non-member counterfactual-distance distributions for the three counterfactual explanation methods.
    (a) NICE on Location,
    (b) Wachter on Purchase-100, and
    (c) Growing Spheres on Texas-100.}
    \label{fig:cf-distance-distributions}
\end{figure}

Figure~\ref{fig:cf-distance-distributions} presents member and non-member counterfactual-distance distributions, which are detailed in Section~\ref{sec::tabular_cf}.

\subsection{Second Stage DTC Ablation Using a Joint Objective}
\label{app:dtc-ablation}
We provide an ablation study of the second-stage distance-threshold calibration objective in Table~\ref{tab:threshold_calibration_ablation} which facilitates the discussion in Section~\ref{sec::tabular_cf}. The objective maximizes balanced accuracy instead of minimizing the difference between member and nonmember accuracy. 

\begin{table*}[t]
\centering
\caption{Ablation of the second-stage distance-threshold calibration objective. 
``Gap'' is the original objective that minimizes the member/non-member calibration-accuracy difference; 
``Joint'' maximizes their balanced accuracy. 
The final column reports the absolute held-out difference between member and non-member accuracies. 
Directions indicate the membership decision rule.}
\label{tab:threshold_calibration_ablation}
\scriptsize
\setlength{\tabcolsep}{4pt}
\resizebox{\textwidth}{!}{%
\begin{tabular}{lllrrrr}
\toprule
Method & Dataset (distance) & Threshold: Gap $\rightarrow$ Joint &
ASR$_{\mathrm{Gap}}$ (\%) & ASR$_{\mathrm{Joint}}$ (\%) &
$\Delta$ ASR (pp) &
$|\mathrm{Acc}_{\mathrm{mem}}-\mathrm{Acc}_{\mathrm{non}}|$ \\
\midrule
\multicolumn{7}{l}{\textit{Image attacks}} \\
SEDC & CIFAR-10 (segment count)  & $>2 \rightarrow >2$ & 72.17 & 72.17 & \phantom{+}0.00 & 9.95 $\rightarrow$ 9.95 \\
SEDC & CIFAR-100 (segment count) & $>2 \rightarrow >1$ & 65.54 & 68.42 & $+2.88$ & 9.70 $\rightarrow$ 52.96 \\
DiVE & CIFAR-10 (LPIPS)           & $\leq 0.33994 \rightarrow \leq 0.30868$ & 84.27 & 85.87 & $+1.61$ & 6.47 $\rightarrow$ 11.93 \\
DiVE & CIFAR-100 (SSIM)           & $\leq 0.37663 \rightarrow \leq 0.38916$ & 75.33 & 75.08 & $-0.25$ & 2.02 $\rightarrow$ 11.63 \\
\midrule
\multicolumn{7}{l}{\textit{Tabular attacks}} \\
NICE & Purchase-100 & $>6.1635 \rightarrow >5.5554$ & 84.79 & 87.21 & $+2.43$ & 3.00 $\rightarrow$ 13.00 \\
NICE & Location     & $>8.1828 \rightarrow >7.9683$ & 73.97 & 75.16 & $+1.19$ & 2.86 $\rightarrow$ 3.02 \\
NICE & Texas-100    & $>22.5220 \rightarrow >15.1152$ & 62.71 & 62.29 & $-0.43$ & 0.00 $\rightarrow$ 34.57 \\
\addlinespace
Growing Spheres & Purchase-100 & $>6.3567 \rightarrow >5.3271$ & 95.00 & 94.43 & $-0.57$ & 2.57 $\rightarrow$ 10.86 \\
Growing Spheres & Location     & $>7.4628 \rightarrow >7.6169$ & 80.16 & 80.40 & $+0.24$ & 9.52 $\rightarrow$ 8.10 \\
Growing Spheres & Texas-100    & $>23.6475 \rightarrow >19.3501$ & 82.00 & 83.64 & $+1.64$ & 8.00 $\rightarrow$ 23.29 \\
\addlinespace
Wachter & Purchase-100 & $>1.9382 \rightarrow >1.7404$ & 92.14 & 94.93 & $+2.79$ & 0.57 $\rightarrow$ 8.71 \\
Wachter & Location     & $>2.6870 \rightarrow >2.6861$ & 79.52 & 79.52 & \phantom{+}0.00 & 3.81 $\rightarrow$ 3.81 \\
Wachter & Texas-100    & $>2.8330 \rightarrow >2.0718$ & 76.29 & 79.21 & $+2.93$ & 6.86 $\rightarrow$ 28.43 \\
\bottomrule
\end{tabular}%
}
\end{table*}

\section{Comparison to Related Work}\label{app:comparison}
Table~\ref{tab:comparison} presents a comparison of prior MIA work on Purchase-100, Location, and Texas-100, which facilitates the discussion in Section~\ref{sec:comparison}.

\begin{table*}[t]
    \centering
    \caption{Comparison of MIAs on Purchase-100, Location, and Texas-100.
    ASR and AUC are reported as fractions; ``--'' denotes an unreported result.
    Queries count attacker-issued target-model or explanation-service interactions.}
    \label{tab:comparison}

    \scriptsize
    \setlength{\tabcolsep}{1.8pt}
    \renewcommand{\arraystretch}{1.15}

    \resizebox{\textwidth}{!}{%
    \begin{tabular}{
        @{}
        l
        p{2.55cm}
        p{3.15cm}
        cc
        cc
        cc
        cc
        @{}
    }
        \toprule
        \textbf{Method} &
        \textbf{Attack signal} &
        \textbf{Experimental setup} &
        \multicolumn{2}{c}{\textbf{Queries}} &
        \multicolumn{2}{c}{\textbf{Purchase}} &
        \multicolumn{2}{c}{\textbf{Location}} &
        \multicolumn{2}{c}{\textbf{Texas}} \\
        \cmidrule(lr){4-5}
        \cmidrule(lr){6-7}
        \cmidrule(lr){8-9}
        \cmidrule(lr){10-11}

        & & &
        \scriptsize One-time &
        \scriptsize Per candidate &
        \scriptsize ASR & \scriptsize AUC &
        \scriptsize ASR & \scriptsize AUC &
        \scriptsize ASR & \scriptsize AUC \\
        \midrule

        Gap Attack~\cite{yeom-overfitting}
        & Prediction correctness
        & Evaluated on our target models; no calibration
        & 0
        & 1 label
        & .692 & --
        & .661 & --
        & .652 & -- \\

        Choquette-Choo et al.~\cite{choo-label-only}
        & Robustness to input perturbations
        & Shadow/source model for threshold selection; hard-label access
        & 0 target
        & $10^4$ labels
        & .874 & --
        & \textbf{.892} & --
        & .803 & -- \\

        Shokri et al.~\cite{shokri-xai}
        & Variance of LIME coefficients
        & Optimal variance threshold; Purchase and Texas only
        & --
        & 1 expl.$^a$
        & $\leq .500^a$ & --
        & -- & --
        & $\leq .500^a$ & -- \\

        Ye et al.~\cite{ye_classical_mia} (Attack P)
        & Population-calibrated loss score
        & 1,000 population records per class; score access
        & $10^5$ scores$^b$
        & 1 score
        & -- & .818
        & -- & --
        & -- & -- \\

        YOQO~\cite{yoqo-mia}
        & Shadow-crafted membership query
        & Online: 16 in/out pairs; offline: 16 out-models
        & 0 target
        & 1 label
        & $.6828/.700^c$ & --
        & $.775/.8325^c$ & --
        & -- & -- \\

        DHAttack~\cite{li_enhanced_2025}
        & Fixed-direction boundary distance
        & 256 shadow models; 40,000 reference records
        & $4{\times}10^4$ labels$^d$
        & $\leq 30$ labels
        & -- & .727
        & -- & --
        & -- & -- \\

        LoMime~\cite{lomime}
        & Boundary distance on an extracted surrogate
        & Label-only extraction followed by offline MIA
        & $10^6/10^5/10^6{}^e$
        & 0 target
        & .8495 & .8530
        & .8563 & .8582
        & .7823 & .7996 \\

        \midrule

        \textbf{\sys{} (LIME)}
        & Local boundary distance
        & Calibrated threshold and direction; no shadows
        & $30$ expl.$^f$
        & 1 expl.
        & .850 & .914
        & .710 & .780
        & .640 & .688 \\

        \textbf{\sys{} (NICE)}
        & Counterfactual distance
        & NICE counterfactual; no shadows
        & $30$ expl.$^f$
        & 1 expl.
        & .848 & .916
        & .740 & .812
        & .627 & .676 \\

        \textbf{\sys{} (Growing Spheres)}
        & Counterfactual distance
        & Growing Spheres counterfactual; no shadows
        & $30$ expl.$^f$
        & 1 expl.
        & \textbf{.950} & \textbf{.976}
        & .802 & \textbf{.871}
        & \textbf{.820} & \textbf{.881} \\

        \textbf{\sys{} (Wachter)}
        & Counterfactual distance
        & Wachter counterfactual; no shadows
        & $30$ expl.$^f$
        & 1 expl.
        & .921 & .967
        & .795 & .847
        & .763 & .798 \\

        \bottomrule
    \end{tabular}%
    }

    \vspace{2pt}
    \begin{minipage}{0.99\textwidth}
        \scriptsize
        $^a$ Shokri et al. do not report exact ASRs, calibration costs, or
        LIME's internal prediction-query count; ``1 expl.'' denotes one
        released explanation per candidate.
        $^b$ Attack P uses 1,000 population records for each of the 100
        Purchase classes, giving $10^5$ target-score evaluations.
        $^c$ Values are YOQO online/offline.
        $^d$ DHAttack's $4{\times}10^4$ queries are for Purchase reference-data
        relabeling; calibrating a fixed binary threshold may require additional
        queries, whereas the reported Purchase result is AUC.
        $^e$ LoMime budgets are one-time extraction queries for
        Purchase/Location/Texas; membership inference is subsequently offline.
        $^f$ XposedMIA uses 30 explanation requests for calibration and one
        externally issued explanation request per candidate; service-side
        model evaluations are not counted separately.
    \end{minipage}
\end{table*}

\section{Implementation Details}

\subsection{XAI Setup}\label{app:implementation-details}
For the explanation tools, we use an authors' implementation whenever it is directly applicable to our setting. LIME explanations are generated using the official Python implementation of LIME~\cite{lime}, specifically, we use the LIME Tabular Explainer. Unless otherwise stated, we use LIME's default configuration, including 5,000 perturbed samples and the Euclidean distance metric. For LIME, each response contains the predicted label together with the local surrogate's weights and intercepts. 

For NICE we use the official \texttt{NICEx} implementation (version~0.2.3). For the remaining methods, a local implementation or adaptation was required to support a common explanation-service interface, identical target-model checkpoints and preprocessing, multiclass prediction, explicit generation failures, and reproducible control of the generation budget. 

Wachter~\cite{counterfactual-watcher} does not provide an official maintained implementation, we therefore implement its original squared target-output loss and inverse-MAD-weighted $L_1$ proximity objective. We use Adam as the numerical optimizer and extend target selection to multiclass models by optimizing toward the most probable alternative classes. Candidate values are constrained to the feature ranges observed in the provider's training data, and the closest valid candidate is returned.

Although reference code is available for Growing Spheres, we reimplement its two defining stages to integrate it with the same TensorFlow target models, preprocessing pipeline, batched provider-side inference, deterministic random
seeding, and explicit failure and budget semantics used by the other explainers. The implementation retains the published algorithmic structure: volume-uniform sampling within expanding Euclidean shells, selection of the nearest differently classified candidate, and subsequent feature sparsification. The search is untargeted and accepts any label change, thereby supporting the multiclass datasets considered in our evaluation. Feasibility clipping is disabled by default so that it does not alter the original unconstrained search.

For the visual methods, the released implementations are specialized to different experimental settings. The released SEDC workflow is centered on ImageNet-scale inputs and MobileNetV2, whereas our evaluation uses MNIST- and CIFAR-scale images and separately trained multiclass target models. We therefore implement the published segment-removal search using grid or SLIC segmentation and Gaussian-blur replacement, which is a replacement operation supported by the authors' implementation. We additionally introduce explicit evaluation, runtime, queue-size, and segment-count limits and perform a final set-minimal reduction; these choices make failures observable and prevent unbounded provider-side computation.

DiVE's released implementation and pretrained models target different image domains and binary prediction tasks. We adapt the authors' reference implementation to TensorFlow/NHWC, $28\times28$ and $32\times32$ inputs, one-and three-channel images, and multiclass classifiers. We preserve the $\beta$-TCVAE architecture and its reconstruction, latent-sparsity, and diversity objectives, replacing only the binary target loss with multiclass target cross-entropy. DiVE generates multiple candidates internally, after which the provider returns the valid candidate having the smallest published proximity objective. These adaptations concern compatibility rather than the membership decision rule.

\subsection{Dataset Splits Among the Different Attacks}
\label{app:dataset-splits}
We provide a detailed information on dataset splitting and attack-set construction in Table~\ref{tab:dataset_splits}.

\begin{table*}[t]
\centering
\caption{Dataset splitting and attack-set construction for the selected configurations.
$T$ denotes the target-model member-source pool and $N$ the disjoint non-member source pool.
``Fit/validation'' gives the target model's gradient-training and validation counts.
For supervised calibration, $c/i$ denotes the number of calibration and held-out inference examples \emph{per membership group}; thus, each attack uses $2c$ calibration and $2i$ inference queries.
For each image dataset, SEDC and DiVE use the identical fixed target model and member-source pool; only their non-member pool sizes and calibration/inference allocations differ.}
\label{tab:dataset_splits}
\scriptsize
\setlength{\tabcolsep}{4pt}
\resizebox{\textwidth}{!}{%
\begin{tabular}{llllrrl}
\toprule
Attack family & Dataset & Target model and member source & Method &
$|N|$ & $c/i$ per group & Attack-subset construction \\
\midrule
\multicolumn{7}{l}{\textit{Tabular attacks}} \\
LIME & Purchase-100 &
\makecell[l]{$|T|=6{,}000$\\4,800 / 1,200 fit/validation} &
LIME & 1,000 & 300 / 700 & Balanced \\
LIME & Location &
\makecell[l]{$|T|=1{,}200$\\960 / 240 fit/validation} &
LIME & 900 & 270 / 630 & Balanced \\
LIME & Texas-100 &
\makecell[l]{$|T|=10{,}000$\\8,000 / 2,000 fit/validation} &
LIME & 1,000 & 300 / 700 & Not class-balanced \\
\addlinespace
Counterfactual &
Purchase-100 &
\makecell[l]{$|T|=10{,}000$\\8,000 / 2,000 fit/validation} &
\makecell[l]{NICE, Wachter,\\Growing Spheres} &
1,000 & 300 / 700 & Not class-balanced \\
Counterfactual &
Location &
\makecell[l]{$|T|=1{,}200$\\960 / 240 fit/validation} &
\makecell[l]{NICE, Wachter,\\Growing Spheres} &
1,200 & 270 / 630 & Balanced \\
Counterfactual &
Texas-100 &
\makecell[l]{$|T|=10{,}000$\\8,000 / 2,000 fit/validation} &
\makecell[l]{NICE, Wachter,\\Growing Spheres} &
1,000 & 300 / 700 & Balanced \\
Counterfactual &
Adult &
\makecell[l]{$|T|=10{,}000$\\8,000 / 2,000 fit/validation} &
\makecell[l]{NICE, Wachter,\\Growing Spheres} &
1,000 & 300 / 700 & Balanced \\
\midrule
\multicolumn{7}{l}{\textit{Image attacks: shared target model within each dataset}} \\
\multirow{2}{*}{Image counterfactual} &
\multirow{2}{*}{CIFAR-10} &
\multirow{2}{*}{\makecell[l]{Shared CNN target\\$|T|=10{,}000$; 9,000 / 1,000 fit/validation}} &
SEDC & 10,000 & 500 / 1,000 & Balanced \\
& & & DiVE & 5,000 & 100 / 1,000 & Balanced \\
\addlinespace
\multirow{2}{*}{Image counterfactual} &
\multirow{2}{*}{CIFAR-100} &
\multirow{2}{*}{\makecell[l]{Shared ResNet-18 target\\$|T|=50{,}000$; 45,000 / 5,000 fit/validation}} &
SEDC & 10,000 & 500 / 1,000 & Balanced \\
& & & DiVE & 10,000 & 100 / 1,000 & Balanced \\
\bottomrule
\end{tabular}%
}

\vspace{2pt}
\parbox{0.96\textwidth}{\footnotesize
\textit{Implementation details.}
All calibration and inference subsets are disjoint. The supervised calibration subsets are used only to select the distance threshold; ASR is computed exclusively on the held-out inference subsets.
For tabular target models, Keras reserves 20\% of $T$ as a validation split during fitting.
For SEDC and DiVE, the 10\% target-validation split is excluded from the image-attack member population.
DiVE additionally uses a separate explainer validation set: for CIFAR-10, its explainer is trained on 8,100 target-fitting images and validated on 810 target-validation images; for CIFAR-100, its explainer uses 40,500 target-fitting images and 4,500 source-holdout validation images, which are excluded from the DiVE member attack population.
}
\end{table*}

\end{document}

%% file: sections/1-introduction.tex
\section{Introduction}
\label{sec:intro}

The rapid adoption of machine learning (ML) models for everyday tasks is increasing their use by individuals, businesses, and organizations~\cite{marketing_mlaas,mlaas_trend,mlaas_organization}. Since ML models are generally resource-intensive, require training with large datasets, and are difficult to deploy, machine learning as a service (MLaaS) platforms are becoming more common~\cite{mlaas}. These platforms provide models that users can query with their own data and receive the output back~\cite{tramer-stealing}. 

A major problem arises in such MLaaS platforms when providers do not want to disclose the internals of their models or datasets, as these are considered intellectual property~\cite{tramer-stealing}. While this protects model internals, the resulting black-box nature introduces risks. Complex models such as neural networks (NNs) offer only the input and output to the user, without revealing what leads to a specific prediction or classification decision~\cite{survey_explaining_black_box}. This problem is especially salient in high-stakes domains, including clinical decision support and credit or financial decision-making, where explanations support clinical oversight, accountability, and responsible deployment~\cite{xai_medical, amann_explainability_2020, bussmann_explainable_2021}. Moreover, specific legal regimes require reason-giving or meaningful explanations for certain automated decisions, including adverse credit actions in the US and decisions based on specified high-risk AI systems in the EU~\cite{eu_ai_act, 12_cfr}. For this reason, some MLaaS platforms offer explanations through explainable artificial intelligence (XAI) tools~\cite{AWS_shap,googleCloud,ibmExp,azureExp}. 

XAI enables users to interpret model predictions and gain insight into the factors influencing their decisions. A wide range of explanation techniques has been developed, each adopting a distinct methodological approach and offering various perspectives on how a model arrives at its predictions~\cite{lime,shap,integrated-gradient,grad-cam,xrai,smoothGrad}. 


While XAI facilitates transparency, the shared information inherently exposes behavioral signals that expand the model's attack surface. ML models are vulnerable to a wide range of security/privacy attacks, including model extraction~\cite{AUTOLYCUS,modelExtraction_chandra,modelExtraction}, adversarial attacks~\cite{ghorbani}, model inversion~\cite{model-inversion-fredrikson,modelInversion}, and membership inference attacks (MIAs)~\cite{shokri-mia,ye_classical_mia,oslo,sersem}. Prior work has demonstrated that the integration of XAI techniques can further increase the vulnerability of ML models to such threats, effectively broadening the attack surface~\cite{shokri-xai,AUTOLYCUS,ezzeddine,lime-shap-fooling,modelInversion,modelConstruction,modellabel-only_xai}. A successful MIA can lead to serious data leakage, thereby compromising data privacy. For example, if a model is trained to classify a sensitive property such as cancer type, a successful MIA could reveal that a specific individual was included in the training set, exposing sensitive information about their health. Moreover, since training datasets may be considered intellectual property, a successful MIA on a public dataset could reveal whether the dataset was used to train the model.

\begin{figure*}[h]
    \centering
    \includegraphics[width=0.98\textwidth]{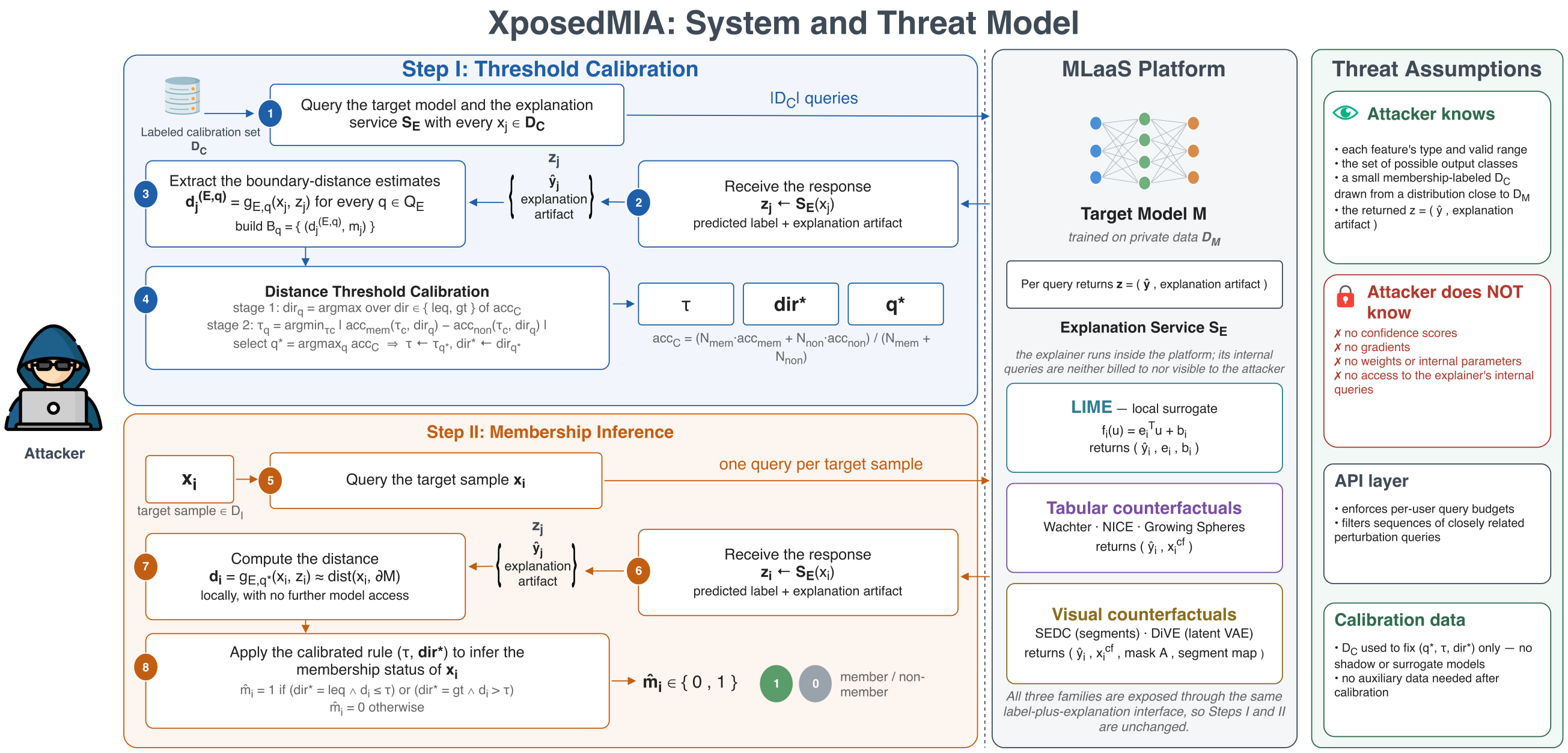}
    \caption{Overview of \sys. The attacker interacts with the MLaaS platform only
through the explanation service $\mathcal{S}_\mathcal{E}$, which returns
$z=(\hat{y},\text{explanation artifact})$ under Local Interpretable Model-Agnostic Explanations (LIME)~\cite{lime}, tabular counterfactuals (Wachter~\cite{counterfactual-watcher}, Nearest Instance Counterfactual Explanations
(NICE)~\cite{nice-counterfactual}, and Growing Spheres (GS)~\cite{GS-counterfactual}), or visual counterfactuals
(Search for EviDence Counterfactual (SEDC)~\cite{vermeire-sedc-image} and Diverse Valuable Explanations
(DiVE)~\cite{dive-rodriguez}). In \textbf{Step I},
the attacker queries $\mathcal{S}_\mathcal{E}$ with the labeled calibration set
$\mathcal{D}_C$, converts each response into a boundary-distance estimate
$d^{(\mathcal{E},q)}_j=g_{\mathcal{E},q}(x_j,z_j)$, and fixes
$(q^{*},\tau,\mathit{dir}^{*})$ via Algorithm~\ref{alg:calibration}
(Appendix~\ref{app:dtc}). In \textbf{Step II}, one query per target sample $x_i$
yields $z_i$, from which $d_i=g_{\mathcal{E},q^{*}}(x_i,z_i)$ is computed locally
and compared against $(\tau,\mathit{dir}^{*})$ (Algorithm~\ref{alg:pipeline}).
Only the extractor $g_{\mathcal{E},q}$ is explanation-specific
(Sec.~\ref{sec:distance}); both steps are otherwise identical across explainers.}
    \label{fig:method_overview}
\end{figure*}

In this work, we study whether explanations can be turned into a membership signal in a black-box setting where the adversary receives a predicted label and its accompanying explanation. We present \sys (illustrated in Figure~\ref{fig:method_overview}), a one-shot explanation-based MIA framework that derives a proxy for the queried sample's distance to the model decision boundary from the returned explanation. This estimated boundary distance is then used to infer the membership status. Prior label-only boundary-distance attacks~\cite{choo-label-only,li_enhanced_2025,li-label-only} require a substantial number of queries or shadow model training with large datasets to estimate the position of the decision boundary and do not leverage information from XAI methods. As a result, these approaches incur high query and computational costs, limiting their practicality in real-world black-box settings. In contrast, \sys reduces the attack cost to a single external service query per target sample after a one-time calibration phase. 
Existing \textit{XAI-based attacks}, on the other hand, typically rely on stronger adversarial assumptions or incur higher attack costs, such as white-box model access, shadow-model training, or repeated queries to the target model~\cite{liu-xai,shokri-xai}. To the best of our knowledge, no prior work has explicitly used LIME's local linear surrogate as a proxy for decision-boundary distance under a black-box setting or investigated visual counterfactual explanations for membership inference on image classifiers.


In addition, \sys is designed to minimize the number of queries, which is crucial in MLaaS settings where application programming interfaces (APIs) often enforce query budgets and detect adaptive query patterns. Existing perturbation-based attacks require multiple queries per sample, increasing their visibility and making them easier to detect. In contrast, \sys uses only a single standard query per target instance after calibration. We thus demonstrate that exposing explanations alongside model predictions introduces a lower-interaction attack surface for MIA, enabling more stealthy attacks under practical API constraints.

\sys consists of two main steps. First, we calibrate the distance threshold $\tau$ by issuing queries to the MLaaS platform and collecting predicted labels together with their corresponding explanations. This calibration phase is used only to determine a membership decision threshold; it does not require training shadow models. While the best performance is achieved with approximately 100 calibration queries, the method remains effective with smaller calibration sets as few as 30 samples. Then, the attack requires only a single query per target sample and makes its decision based on the sample’s distance to the estimated decision boundary. This enables a one-shot, explanation-guided black-box MIA that leverages explanations without requiring shadow-model training.

We note here that the calibration requirement is substantially weaker than the data requirements of shadow-model-based pipelines. The calibration set is used only to determine a threshold and decision direction, rather than to train surrogate models or approximate the target model. On the other hand, prior attacks require significantly larger auxiliary datasets to train models that mimic the target, sometimes reaching up to $10^6$ queries, resulting in much heavier data requirements. We summarize our main contributions below:
\begin{itemize}
    \item We demonstrate that releasing LIME and counterfactual explanations with model outputs can introduce measurable membership leakage in a label- and explanation-only black-box setting.
    \item We introduce the first unified \textbf{one-shot} explanation-based membership inference framework that leverages explanations as the attack surface: a single query suffices to recover membership information about a target sample.
    \item We evaluate \sys across four tabular and two image benchmark datasets. Our experimental results, ablation studies, and comparisons with prior work demonstrate its strong effectiveness and efficiency.
    \item We propose and evaluate, to the best of our knowledge, the first extension of MIA to visual counterfactual explanations on image classifiers using SEDC~\cite{vermeire-sedc-image} and DiVE~\cite{dive-rodriguez}.

\end{itemize}

\par\noindent\textbf{Scope.} In this work, we base our MIA method on LIME~\cite{lime} as a local surrogate explanation method and extend our method to other XAI methods, including three tabular counterfactual methods, Wachter~\cite{counterfactual-watcher}, NICE~\cite{nice-counterfactual}, and Growing Spheres~\cite{GS-counterfactual}, and two visual counterfactual methods, SEDC~\cite{vermeire-sedc-image} and DiVE~\cite{dive-rodriguez}. These methods span local surrogate and counterfactual explanations, providing different forms of information about the model's decision structure. Local surrogate explanations approximate the model's behavior around the queried sample, while counterfactual explanations identify changes to the input that alter the model's prediction. We derive a proxy for the queried sample's distance to the decision boundary from explanation outputs. More details on these XAI methods and the rationale behind targeting them can be found in Section~\ref{sec:background}.


%% file: sections/2-related_work.tex
\section{Related Work}
\label{sec:related-work}

We organize prior work on MIAs into five categories: (i) \emph{classical MIAs}, which typically rely on confidence scores, loss values, or internal model signals; (ii) \emph{label-only MIAs}, which operate in black-box settings using only the predicted label; (iii) \emph{low-query label-only MIAs}, which aim to reduce the substantial query cost of earlier label-only attacks; (iv) \emph{feature-attribution-based MIAs}, which exploit information exposed by explanation methods that return feature attributions; and (v) \emph{counterfactual-explanation-based attacks}, which exploit information revealed by counterfactual explanations for membership inference, model extraction, or reconstruction.

\par\noindent\textbf{Classical Membership Inference Attacks.} Classical MIAs are commonly studied in black-box or white-box settings in which the attacker has access to confidence scores or other signals produced by the target model. In black-box classification settings, these scores typically correspond to posterior or confidence vectors over the output classes. Shokri et al.~\cite{shokri-mia} introduced the first MIA against machine-learning classifiers, training shadow models on auxiliary data drawn from the same distribution as the target model's training data. The shadow models are queried with member and non-member samples, and their outputs are subsequently used to train a separate membership classifier. Salem et al.~\cite{salem2018-model_data_independent} later showed that comparable attacks can be constructed using only a single shadow model.

More recently, Ye et al.~\cite{ye_classical_mia} formulated membership inference as a hypothesis-testing problem and developed stronger loss-based attacks using reference models. Carlini et al.~\cite{carlini_membership_2022} further demonstrated that average-case metrics can obscure privacy leakage in operationally relevant regimes and advocated evaluating MIAs at low false-positive rates (FPRs). They introduced the likelihood ratio attack (LiRA), which has since become an important benchmark.

Nasr et al.~\cite{nasr_whitebox_mia} extended MIA to the white-box setting and showed that access to internal signals, including gradients and intermediate representations, can further amplify membership leakage. Their results demonstrate that stronger access to the target model can reveal membership information beyond that available from prediction outputs alone. These attacks generally rely on confidence or loss information, internal model signals, or auxiliary model training. In contrast, \sys requires none of these forms of access.

\par\noindent\textbf{Label-only Membership Inference Attacks.} Label-only MIAs consider a more restrictive black-box setting in which the attacker observes only the predicted class label. Yeom et al.~\cite{yeom-overfitting} proposed a simple label-only baseline in which a correctly classified sample is predicted to be a member. This attack, commonly referred to as the \emph{gap attack}, is directly related to the difference between the target model's training and test accuracy and is widely used as a baseline in label-only settings.

Choquette-Choo et al.~\cite{choo-label-only} introduced a stronger label-only MIA based on a sample's distance to the decision boundary. The intuition is that training samples may lie farther from the decision boundary and therefore require larger perturbations to change their predicted label. They estimate this distance using an adversarial sample generation technique called HopSkipJump~\cite{hopskipjump}, which searches for a small perturbation that changes the target model's prediction. The magnitude of the resulting perturbation is then used as a proxy for boundary distance and, consequently, as a membership signal.

Li et al.~\cite{li-label-only} proposed a label-only MIA based on transferability between shadow and target models. They additionally introduced boundary-distance attacks based on HopSkipJump~\cite{hopskipjump} and the query-efficient boundary-based black-box attack~\cite{qeba}, removing the dependence on this transferability assumption. More recently, Li et al.~\cite{li_enhanced_2025} proposed an enhanced label-only MIA that learns boundary-distance statistics from shadow models and performs MIA using $\sim10-50$ target-model queries per sample.

A common limitation of these MIAs is therefore their per-sample query cost, since estimating the decision boundary generally requires repeatedly querying the target model. Such repeated probing may be constrained by rate limits or abuse-detection mechanisms in MLaaS environments. \sys instead performs a one-time calibration using a small calibration set and subsequently requires only one explanation request per target sample.

sec:low-query-label-only{Low-query Label-only MIAs.} Peng et al.~\cite{oslo} proposed OSLO, a one-shot label-only MIA based on transferable adversarial examples. OSLO requires only a single hard-label query to the target model for each inference sample; however, this inference relies on an offline stage in which the attacker trains an ensemble of surrogate models on an auxiliary dataset and uses these models to construct sample-specific adversarial examples. In their CIFAR experiments, OSLO uses 25,000 auxiliary samples to train its surrogate models. Oksuz et al.~\cite{lomime} introduced LoMime, a label-only MIA that reduces the amortized query cost of membership inference through model extraction. LoMime first extracts a surrogate of the target model using hard-label queries and then performs MIA locally on the extracted model, requiring no additional target-model queries for subsequent inference samples. In its strongest configurations, LoMime uses up to $10^6$ target-model queries during this one-time extraction stage, making the approach particularly attractive when this cost can be amortized over a large number of membership queries. Overall, OSLO and LoMime reduce the per-sample query cost by shifting substantial work to an initial surrogate-training or model-extraction. Although this amortization is advantageous when attacking very large numbers of samples, the corresponding upfront data, computation, and target-query requirements can be substantial and may be constrained by MLaaS. In contrast, \sys achieves one-shot inference without surrogate-model training, adversarial-example construction, or model extraction: it requires only a small calibration set to fix a scalar threshold and its direction, followed by a single explanation request per target sample. Thus, \sys has a more realistic assumption about the amount of auxiliary data.

\par\noindent\textbf{Explanation-Based Membership Inference Attacks.} A separate line of work considers MLaaS settings in which an explanation is returned alongside the model's prediction and investigates whether the additional information leaks membership.

Shokri et al.~\cite{shokri-xai} studied MIAs in which the adversary obtains both prediction outputs and feature-based explanations, including Integrated Gradients (IG)~\cite{integrated-gradient}, Layer-wise Relevance Propagation (LRP)~\cite{lrp}, LIME~\cite{lime}, and SmoothGrad~\cite{smoothGrad}. They showed that statistics derived from predictions and explanations, including their variance, can reveal membership information. Their attacks use either an optimal threshold, which assumes knowledge unavailable to a practical attacker, or thresholds calibrated using shadow models trained on auxiliary data. Their approach is effective for several backpropagation-based explanation methods, including gradients, IG, and LRP, but does not obtain useful membership signals from perturbation-based methods such as LIME and SmoothGrad. 

Liu et al.~\cite{liu-xai} further showed that feature-attribution explanations can expose robustness differences between members and non-members. Their attack iteratively perturbs the most and least important pixels identified by an attribution map and uses the resulting model behavior as membership features. The attack requires training auxiliary models to approximate the target model's behavior, and additionally relies on posterior outputs. Their approach is primarily evaluated on image classifiers and achieves strong attack performance, but requires substantially richer attacker-side machinery than a direct threshold-based attack.

Few prior works combine explanation access with hard-label prediction access while avoiding confidence scores and model internals. Ma et al.~\cite{modellabel-only_xai} use explanations to guide iterative feature removal until the target model's predicted label changes. The resulting robustness score is then used for membership inference, with shadow models used to calibrate the decision. Although this approach substantially reduces the query cost compared with earlier label-only attacks, obtaining the membership signal still requires repeatedly querying the target model and incurs hundreds of queries per target sample in addition to shadow-model training.

Overall, prior feature-attribution-based MIAs commonly rely on confidence or posterior information, shadow-model training, or repeated target-model queries. \sys instead derives its membership signal directly from the returned explanation and requires only lightweight calibration before one-shot inference.

\par\noindent\textbf{Counterfactual-Explanation-Based Attacks.} Counterfactual explanations expose a different form of information about a model's local decision. Given an input $x$, a counterfactual explanation produces a modified instance $x'$ whose prediction differs from that of $x$ while seeking to keep the modification small according to a notion of proximity~\cite{counterfactual-watcher,GS-counterfactual,nice-counterfactual}. Consequently, the distance $\lVert x-x' \rVert$ provides an explanation-derived measure of the change required to alter the model's decision.

This property has been exploited for model extraction and reconstruction. A\"ivodji et al.~\cite{modelExtraction} query a target service for predictions and counterfactual explanations and use the resulting samples to train a surrogate model. Wang et al.~\cite{wang2022dualcf} improve the efficiency of this strategy by additionally obtaining a counterfactual of each returned counterfactual, reducing the decision-boundary shift. Dissanayake and Dutta~\cite{dissanayake2024polytope} formalize this intuition by showing that a sufficient number of optimal counterfactuals can be used to approximate convex decision boundaries. Related reconstruction attacks have also been demonstrated for feature-attribution explanations: Milli et al.~\cite{modelConstruction} reconstruct models from gradient-based explanations, while Oksuz et al.~\cite{AUTOLYCUS} extract models from LIME and SHAP outputs.


Counterfactual explanations have also been shown to support membership inference. Pawelczyk et al.~\cite{pawelczyk2023recourse} use the recourse cost $c(x,x')$ as a membership signal, and the intuition is that training can move the decision boundary away from member samples, causing members and non-members to exhibit systematically different counterfactual distances. They instantiate this idea through two attacks: a direct counterfactual-distance threshold, which sweeps possible values to select an optimal threshold, and a likelihood-ratio test inspired by Carlini et al.~\cite{carlini_membership_2022} that uses auxiliary data and shadow models to estimate reference distributions. They further observe that the direction of the membership signal can depend on the counterfactual generator, with some methods producing smaller rather than larger distances for members.

Ezzeddine et al.~\cite{ezzeddine-2026} further demonstrated that counterfactual explanations can increase membership leakage. Their approach augments a shadow-model-based MIA with information derived from counterfactual explanations and evaluates the attack on two structured numerical datasets. On the EEG dataset, incorporating counterfactual information increases the reported attack success rate from approximately $58\%$ to $77\%$. Unlike their learned shadow-model-based attack, \sys performs MIA using a lightweight calibrated distance rule and requires no shadow-model training.

Taken together, prior work establishes that both feature-attribution and counterfactual explanations can expose membership-relevant information. \sys extends this line of work in three directions: it exploits LIME's local surrogate geometry as a membership signal, performs inference without confidence vectors, model internals, or shadow-model training, and, to the best of our knowledge, provides the first systematic MIA exploiting visual counterfactual explanations for image classifiers.

%% file: sections/3-background.tex
\section{Explainable Artificial Intelligence (XAI)}
\label{sec:background}

Explainable Artificial Intelligence (XAI) aims to make the outputs of complex ML models more interpretable~\cite{LONGO2024102301}. Since modern models, particularly NNs, often provide limited insight into the reasoning behind their predictions, some MLaaS providers expose post-hoc explanation tools alongside model predictions. For instance, Amazon SageMaker Clarify~\cite{awsClarify} provides SHapley Additive exPlanations (SHAP)-based explanations for model predictions~\cite{AWS_shap}. Post-hoc explanation methods can also be model-agnostic: LIME~\cite{lime} and model-agnostic variants of SHAP~\cite{shap}, for example, infer feature contributions from model outputs without requiring the user to access the model's internal parameters. In this work, we consider two forms of local explanation: local surrogate explanations and counterfactual explanations. We note that we do not consider SHAP, as it lacks the distance-related features that we employ in our attack. 

\descr{Local Interpretable Model-Agnostic Explanations (LIME)~\cite{lime}.}
LIME explains an individual prediction by learning an interpretable approximation of the model's behavior in the neighborhood of the input being explained. Given an input $x$ and a model $\mathcal{M}$, LIME generates perturbed instances around $x$ and obtains the corresponding predictions from $\mathcal{M}$. These samples are weighted according to their proximity to $x$ and used to fit a local interpretable surrogate model, commonly a sparse linear model. The coefficients of this surrogate represent the local contribution of the input features to the prediction and can be presented as feature-importance scores or visual heat maps. Because the surrogate approximates $\mathcal{M}$ only locally, its decision boundary need not coincide exactly with that of the underlying model, particularly when $\mathcal{M}$ is highly nonlinear in the neighborhood of $x$. Nevertheless, LIME is post-hoc, model-agnostic, instance- and prediction-specific, making it applicable to a broad range of black-box prediction services.

\descr{Counterfactual Explanations.}
A counterfactual explanation identifies a modified version of an input that changes the model's prediction while remaining close to the original instance according to a chosen notion of proximity~\cite{counterfactual-watcher}. A common formulation considers an input $x$ and a model $\mathcal{M}$ and searches for a counterfactual $x_{\mathrm{cf}}$ such that $\mathcal{M}(x_{\mathrm{cf}}) \neq \mathcal{M}(x)$. It does this while minimizing, or otherwise constraining, a distance $d(x,x_{\mathrm{cf}})$. Practical counterfactual generators differ in how this objective is formulated and do not necessarily guarantee a globally minimal counterfactual. Unlike feature-attribution explanations, which characterize the contribution of features to the current prediction, counterfactual explanations identify changes that are sufficient to alter the model's decision~\cite{fernandez-counterfactual}. The form of these changes depends on the input domain. For tabular data, counterfactuals typically modify the values of one or more features. For images, visual counterfactual explanations may instead remove or replace image regions, or generate semantically modified images through learned representations~\cite{vermeire-sedc-image,goyal-visual-counterfactual}.

\descr{Tabular Counterfactual Explanations.}
Growing Spheres~\cite{GS-counterfactual} follows a model-agnostic search procedure that samples progressively expanding regions around the factual instance until it identifies an instance with a different prediction, after which the solution can be further sparsified. Wachter et al.~\cite{counterfactual-watcher}, in contrast, formulate counterfactual generation as an optimization problem that balances achieving a desired model output against remaining close to the factual instance. NICE~\cite{nice-counterfactual} follows an example-based approach, beginning from a nearby instance with a different prediction and iteratively replacing feature values to construct a counterfactual. These methods therefore represent different strategies for obtaining decision-changing instances, including search-based, optimization-based, and example-based approaches.

\descr{Search for EviDence Counterfactual (SEDC).}
SEDC builds on the document-explanation approach introduced by Martens and Provost~\cite{orginal-sedc-martens-provost}, in which an explanation consists of a small set of present terms whose removal changes the predicted class. Vermeire et al.~\cite{vermeire-sedc-image} subsequently adapted this approach to image classification and refer to the resulting method as \emph{Search for EviDence Counterfactual} (SEDC). Image SEDC operates on image segments or superpixels. The image is first partitioned into regions, after which SEDC searches for a set of segments whose removal or replacement changes the predicted class. The method supports different segmentation and replacement strategies; Vermeire et al. evaluate several alternatives and use Gaussian blurring as the segment-removal mechanism in their main experiments. The identified decision-changing regions constitute the evidence counterfactual (EdC), while applying these changes to the input produces the corresponding counterfactual image. Standard SEDC searches for a change to any alternative class, whereas SEDC-T extends the approach by allowing a specific target counterfactual class to be specified. In this work, we consider untargeted SEDC, since any valid change in the predicted class provides the decision-changing counterfactual required by our attack.

\descr{Diverse Valuable Explanations (DiVE)~\cite{dive-rodriguez}.}
While SEDC constructs visual counterfactuals by modifying discrete image regions, DiVE generates counterfactual images through a learned latent representation. DiVE first maps an input image into a disentangled latent space using a variational autoencoder (VAE) and then searches for perturbations of the latent representation that change the model's prediction while encouraging the resulting image to remain close to the original. The method jointly optimizes multiple latent perturbations and incorporates a diversity objective so that the resulting counterfactuals capture different decision-changing variations. DiVE additionally proposes variants that restrict highly influential latent factors to discourage trivial counterfactuals and expose alternative attributes. Thus, DiVE provides a complementary visual counterfactual mechanism to SEDC: rather than removing predefined image segments, it produces decision-changing images through modifications in a learned generative representation.


%% file: sections/4-threat.tex
\section{System and Threat Model}
\label{sec:threat}
We consider a black-box MLaaS platform akin to commercial platforms such as Amazon Web Services (AWS) SageMaker Clarify, Google Vertex AI, Azure Machine Learning, or IBM Watson OpenScale~\cite{awsClarify,googleCloud,azureExp,ibmExp}. A target model $\mathcal{M}$ and an explainability tool $\mathcal{E}$ are provided by the MLaaS for user queries. 
We denote the training set as $D_M$, which is used by the model owner to train $\mathcal{M}$. A user can query the service with a sample $x_i$ to obtain the corresponding predicted class $\hat{y}_i$ and its explanation $z_i$. Users only have access to the predicted labels and explanations; the corresponding confidence scores and gradient information are not disclosed. We refer to this access model as an explanation-based black-box setting. We consider an API layer that enforces per-user query budgets and filters sequences of closely related perturbation queries.

The attacker has no access to the model’s weights or internal parameters, but is assumed to have knowledge of the feature space, including each feature's type, valid range, and the set of possible output classes. The attacker has access to a small membership-labeled dataset $\mathcal{D}_C$ (see Section~\ref{sec:calibration}), drawn from a distribution close to that of $D_M$, which is only used for the threshold $\tau$ calibration. In particular, $\mathcal{D}_C$ is not used to train shadow or surrogate models. No additional auxiliary data are required after calibration. The attacker could get access to $\mathcal{D}_C$ in many ways. For example if the attacker is participating in training a shared model, they would then have a subset of the training set with high confidence labels. Another way is if a few samples from the training dataset are leaked to the public either through a targeted attack or some internal leak. Given an inference dataset $\mathcal{D}_I$, the attacker's objective is to determine, for each sample $x_i \in D_I$, whether $x_i \in D_M$. 



%% file: sections/5-method.tex
\section{\sys's Overview}
\label{sec:method}

Our attack, \sys, is a unified framework based on the boundary-distance intuition of~\cite{choo-label-only}. While their method estimates distance to the target model's decision boundary through iterative perturbation queries, \sys uses explanations returned by the MLaaS service (see Section~\ref{sec:threat}) as a proxy for boundary information. It derives a boundary-distance score from each explanation and uses it to infer membership. Thus, \sys is a \textbf{one-shot explanation-based black-box MIA framework}: after one-time calibration, each target sample requires only one external service query.

We instantiate the framework with LIME~\cite{lime} and five counterfactual explainers: the tabular methods of Wachter et al.~\cite{counterfactual-watcher}, NICE~\cite{nice-counterfactual}, and Growing Spheres~\cite{GS-counterfactual}; and the visual methods SEDC~\cite{vermeire-sedc-image} and DiVE~\cite{dive-rodriguez}. The framework is shared across all methods; only the conversion of explanations into boundary-distance approximations differs. Calibration and membership decisions remain unchanged.

The attacker neither trains shadow models nor obtains the target model's parameters, gradients, logits, confidence scores, or internal representations. They observe only the model's output and explanation returned by the MLaaS provider. Requiring one query per target after calibration helps evade automatic detection systems and MLaaS rate limits. Consequently, \sys operates through a label-plus-explanation interface: the prediction remains label-only, while the MLaaS response contains both the label and explanation.


\subsection{Attack Intuition}
\label{sec:attack-intuition}
Let $\mathcal{M}$ denote the target classifier and let $\partial\mathcal{M}$ denote its decision boundary. For an explanation method $\mathcal{E}$, the MLaaS explanation service $\mathcal{S}_{\mathcal{E}}$ returns a response $
z_i \leftarrow \mathcal{S}_{\mathcal{E}}(x_i)$ containing the predicted label $\hat{y}_i$ and the explanation produced for $x_i$. Depending on $\mathcal{E}$, the explanation may consist of local surrogate parameters, a counterfactual sample, or a counterfactual image and its associated mask. \sys applies an explanation-specific extractor $g_{\mathcal{E},q}$ to convert this response into a scalar score:
\[
d_i^{(\mathcal{E},q)}
=
g_{\mathcal{E},q}(x_i,z_i)
\approx
\operatorname{dist}\bigl(x_i,\partial\mathcal{M}\bigr),
\]
where $q$ denotes the distance measure used by the extractor. We refer to $d_i^{(\mathcal{E},q)}$ as a \emph{boundary-distance estimate}. For LIME, it is the geometric distance to a locally fitted surrogate boundary. For counterfactual explanations, it is the amount of change between the factual and counterfactual samples. A valid counterfactual lies on the other side of a decision boundary; therefore, the change required to reach it estimates the distance from the factual sample to that boundary. This is generally a proxy rather than the exact minimum distance because an explanation method is not guaranteed to recover the globally closest boundary point.

More generally, a predicted label alone tells in which region the sample $x_i$ is located, while explanations expose additional geometric information in different forms. Thus \sys does not treat the explanation artifact itself as a membership signal, but instead uses the geometric information exposed and transforms it to a scalar boundary distance that can then be used to distinguish between members and non-members. In this sense, explanations provide a shortcut to information about the proximity of a sample to the decision boundary, which would normally require many queries from a prediction-only attacker. We also note that our attack can therefore be affected by the quality of the generated explanations, as incorrect or inaccurate explanations can cause the attack's performance to degrade.

Prior label-only MIAs commonly assume that members lie farther from the decision boundary than non-members~\cite{choo-label-only}. We do not impose this direction universally. Depending on the target model, dataset, explanation generator, and distance measure, members may instead produce smaller estimates. \sys therefore calibrates both the inequality direction and the corresponding decision threshold using a small set of samples with known membership.

\subsection{Estimating the Boundary Distance}
\label{sec:distance}
The attack framework is initially designed for LIME, but is generalized to work with all explanation methods we cover in this work. The only explanation-specific component is the distance scalar $d_i^{(\mathcal{E},q)}$. Once this is obtained, the attack is explanation-agnostic.

\par\noindent{\textbf{LIME}.}
LIME fits a locally weighted interpretable model around the explained sample to approximate the behavior of the target model~\cite{lime}. Consider a tabular sample $x_i\in\mathbb{R}^{n}$ and let $\phi(x_i)$ denote the scaled feature representation used by LIME. The explanation for the predicted class consists of a coefficient vector $e_i\in\mathbb{R}^{n}$ and an intercept $b_i$, defining the local surrogate as
\[
f_i(u)=e_i^{\mathsf{T}}u+b_i.
\]
We use the hyperplane $f_i(u)=t$ as an operational estimate of the local one-versus-rest decision boundary, with $t=0.5$. The resulting LIME boundary-distance estimate is
\[
d_i^{\mathrm{LIME}}
=
\frac{\left|f_i\bigl(\phi(x_i)\bigr)-t\right|}
     {\lVert e_i\rVert_2}
=
\frac{\left|e_i^{\mathsf{T}}\phi(x_i)+b_i-0.5\right|}
     {\lVert e_i\rVert_2}.
\]
Thus, LIME provides an analytical estimate through its local surrogate; \sys does not generate adversarial examples or issue additional attacker-side perturbation queries.

\par\noindent{\textbf{Tabular counterfactual explanations.}}
Given a sample $x_i$, its counterfactual explanation is a sample $x_i^{\mathrm{cf}}$ whose predicted class differs from the original factual sample as:
\[
\mathcal{M}\bigl(x_i^{\mathrm{cf}}\bigr)={\hat{y_i}}, \qquad \mathcal{M}\bigl(x_i)={y_i}, \qquad  {\hat{y_i}}\neq{y_i}.
\]
Given a valid counterfactual, we estimate the boundary distance as $
d_i^{(\mathcal{E},q)}
=
q\bigl(x_i,x_i^{\mathrm{cf}}\bigr)$ where $\mathcal{E}
\in
\{\mathrm{Wachter},\mathrm{NICE},\mathrm{GS}\}$. While this measures the distance to the returned counterfactual rather than directly to the decision boundary, the counterfactual is expected to lie close to the boundary, as it crosses the decision boundary by construction under the assumptions of counterfactual XAI methods. It therefore can be considered a proxy for the distance to the decision boundary.

We evaluate three tabular counterfactual generators. Wachter et al.~\cite{counterfactual-watcher} conduct continuous optimization using a target-output loss and an inverse-median-absolute-deviation-weighted $L_1$ proximity term. It searches over candidate alternative classes and returns the closest counterfactual. NICE~\cite{nice-counterfactual} starts from the nearest unlike instance and optimizes the explanation for sparsity, using a reference pool to construct the counterfactual. Growing Spheres~\cite{GS-counterfactual} samples around the factual sample using an increasingly growing sphere. It keeps iterating until it finds an instance that has a different class than the factual sample.

After obtaining the counterfactual, the attacker computes the distance locally without querying the model $\mathcal{M}$. We measure the distance between the factual and counterfactual samples using standard $L_p$ distances and data-aware measures, including Euclidean, Manhattan, Chebyshev, inverse median absolute deviation (MAD) $L_1$~\cite{counterfactual-watcher}, Mahalanobis, cosine, Canberra, Bray--Curtis, and Gower~\cite{gower-distance} distances. 

\par\noindent{\textbf{SEDC visual counterfactuals.}} 
SEDC constructs a counterfactual by identifying a small set of image segments whose removal changes the predicted class~\cite{vermeire-sedc-image}. Our implementation reproduces one untargeted counterfactual for every request. We first partition the image using a grid (for the MNIST dataset) or Simple Linear Iterative Clustering (SLIC)~\cite{slic-segmentation} (for CIFAR-10 and CIFAR-100). Removing a segment is instantiated by replacing it with a Gaussian-blurred version of itself. It first performs a bounded best-first search over segment subsets and then applies a deterministic reduction step while removing unnecessary segments and maintaining the changed label. We also bound the search by a segment count budget.

The explainer returns the counterfactual image $x_i^{\mathrm{cf}}$, the selected pixel mask, the segment map, and the selected segment identifiers. The attacker does not get access to any of the model's internal parameters that were used during the explanation generation. The boundary distance is then calculated as follows:
\[
d_{\mathrm{seg}}=k,
\qquad
d_{\mathrm{segfrac}}=\frac{k}{K},
\qquad
d_{\mathrm{area}}=\frac{|A|}{HW},
\]
where $k$ is the number of selected segments, $K$ is the total number of segments, $A$ is the selected pixel mask, and $H$ and $W$ are the image dimensions. Note that $d_{\mathrm{segfrac}}$ and $d_{\mathrm{area}}$ generally perform better than $\mathrm{seg}$ since they are normalized.

\par\noindent{\textbf{DiVE visual counterfactuals.}}
DiVE generates diverse visual counterfactuals by optimizing perturbations in the latent space of a generative model~\cite{dive-rodriguez}. DiVE is targeted by nature since its counterfactual loss explicitly optimizes the generated image toward a specific alternative class. For a fixed target class $y_t$, the provider generates a set of candidates satisfying
\[
\mathcal{M}\bigl(x_{i,j}^{\mathrm{cf}}\bigr)=y_t
\neq
\mathcal{M}(x_i).
\]
Among the candidates, the explainer selects one counterfactual sample using DiVE's internal objective that optimizes proximity. This sample is sent to the attacker along with the target class. 

\par\noindent{\textbf{Image distance measures.}}
SEDC and DiVE share the same pixel- and structure-based distances. Let $\Delta_i=\left|x_i-x_i^{\mathrm{cf}}\right|$ for images represented in $[0,1]$. We compute
\begin{align}
d_0
&=
\frac{1}{HW}
\sum_{p=1}^{HW}
\mathbf{1}\!\left[
\max_c \Delta_{i,p,c}>\epsilon
\right],
\label{eq:image-l0}\\
d_1
&=
\operatorname{mean}(\Delta_i),
\label{eq:image-l1}\\
d_2
&=
\sqrt{\operatorname{mean}(\Delta_i^2)},
\label{eq:image-l2}\\
d_{\infty}
&=
\max(\Delta_i),
\label{eq:image-linf}\\
d_{\mathrm{SSIM}}
&=
1-\operatorname{SSIM}\bigl(x_i,x_i^{\mathrm{cf}}\bigr),
\label{eq:image-ssim}
\end{align}
where $\epsilon$ prevents negligible numerical differences from being counted as changed pixels and structural similarity (SSIM) measures structural similarity~\cite{ssim}. SEDC additionally supports the segment- and mask-based measures defined above. As in the tabular setting, every metric for a sample is computed from the same returned counterfactual.

For DiVE, we additionally evaluate Learned Perceptual Image Patch Similarity (LPIPS)~\cite{zhang-lpips}, a perceptual distance based on deep feature representations:
\[
d_{\mathrm{LPIPS}} =
\operatorname{LPIPS}(x_i,x_i^{\mathrm{cf}}).
\]

\subsection{Distance Threshold Calibration (DTC)}
\label{sec:calibration}
This constitutes \textbf{Step I} of \sys as shown in Figure~\ref{fig:method_overview} and summarized in Algorithm~\ref{alg:calibration} in Appendix~\ref{app:dtc}. Prior work suggests that the training members are expected to exhibit a larger distance to the decision boundary compared to non-members~\cite{choo-label-only}. We observe, however, that this assumption does not necessarily hold across all datasets. In certain cases, non-members may exhibit a larger distance. This may be caused by differences in the learned boundary geometry, the distribution and difficulty of member and non-member samples, and the approximation objective of the explanation method. To account for this variability, we use \textbf{a two-stage calibration procedure} that adapts to the relative distance distributions of members and non-members. We assume that the attacker has access to a small calibration dataset $
\mathcal{D}_C=\{(x_j,m_j)\}$, where $m_j\in\{0,1\}$ is the known membership status of $x_j$. We refer to Section~\ref{sec:lime-results} for details on how the size of $\mathcal{D}_C$ affects the results.

For an explanation method $\mathcal{E}$ and a distance measure $q$, the attacker queries each calibration sample and constructs
\[
\mathcal{B}_q
=
\left\{
\bigl(d_j^{(\mathcal{E},q)},m_j\bigr)
\right\},
\]

\noindent using the distance estimates returned by the explainer. Each observed distance is a candidate threshold. For every candidate $\tau_c$ and direction $\mathit{dir}\in\{\mathtt{leq},\mathtt{gt}\}$, it computes the attack accuracy on the calibration dataset to evaluate the effectiveness of that threshold. The direction $\mathtt{leq}$ denotes distances less than or equal to the threshold, while $\mathtt{gt}$ denotes distances greater than the threshold. The sample-weighted overall accuracy is computed as follows:
\[
\mathit{acc}_{C}
=
\frac{
N_{\mathrm{mem}}\mathit{acc}_{\mathrm{mem}}
+
N_{\mathrm{non}}\mathit{acc}_{\mathrm{non}}
}{
N_{\mathrm{mem}}+N_{\mathrm{non}}
}.
\]

This first stage selects the direction $\mathit{dir}_q$ with the highest calibration accuracy $\mathit{acc}_{C}$. The direction determines whether samples closer to the decision boundary should be considered members or non-members. After this direction is determined from the first stage, the second stage is then executed:
\[
\tau_q
=
\underset{\tau_c}{\operatorname{arg\,min}}
\left|
\mathit{acc}_{\mathrm{mem}}(\tau_c,\mathit{dir}_q)
-
\mathit{acc}_{\mathrm{non}}(\tau_c,\mathit{dir}_q)
\right|.
\]

Thus, the threshold equalizes the accuracy for members and non-members. Since several distance measures are available, DTC is executed on each $q$, and the measure with the highest calibration accuracy is selected as the best attack measure.

\subsection{Membership Inference}
\label{sec:method-mia}
This constitutes \textbf{Step II} of \sys (as shown in Figure~\ref{fig:method_overview}). Given an unseen target sample $x_i$, the attacker sends one request to $\mathcal{S}_{\mathcal{E}}$ and obtains its predicted label and explanation. The attacker then computes $
d_i
=
g_{\mathcal{E},q^{*}}(x_i,z_i)
$ without further access to the target model. The calibrated decision rule is
\[
\widehat{m}_i
=
\begin{cases}
1,
&
\mathit{dir}^{*}=\mathtt{leq}
\text{ and }
d_i\leq\tau,
\\
1,
&
\mathit{dir}^{*}=\mathtt{gt}
\text{ and }
d_i>\tau,
\\
0,
&
\text{otherwise},
\end{cases}
\]
where $\widehat{m}_i=1$ denotes a member and $\widehat{m}_i=0$ denotes a non-member.

Algorithm~\ref{alg:pipeline} formalizes our complete end-to-end MIA pipeline. The algorithm is structurally divided into two main phases. The first phase (Lines~1--14) constructs the calibration records $\mathcal{B}_q$, applies DTC independently to each candidate distance measure, and selects the finalized measure $q^{*}$, threshold $\tau$, and direction $\mathit{dir}^{*}$, as detailed in Section~\ref{sec:calibration}. The second phase performs the attack on the inference dataset $\mathcal{D}_I$ (Lines~15--29). For each unseen target sample, the pipeline requests an explanation and computes its boundary-distance estimate using $g_{\mathcal{E},q^{*}}$. It then applies the calibrated decision rule by comparing the distance $d$ with $\tau$ according to $\mathit{dir}^{*}$. Invalid explanation responses are processed using the method-specific failure rule. Finally, the algorithm returns a mapping $\mathcal{R}$ that associates each target sample with its predicted membership status or, where applicable, an abstention.

\begin{algorithm}[t]
    \caption{Unified explanation-distance membership inference pipeline.}
    \label{alg:pipeline}
    \KwIn{Inference dataset $\mathcal{D}_I$; calibration dataset
    $\mathcal{D}_C$ containing tuples $(x,m)$; explanation service
    $\mathcal{S}_{\mathcal{E}}$; candidate distance measures
    $\mathcal{Q}_{\mathcal{E}}$}
    \KwOut{Membership decisions, with abstentions for unsupported image
    counterfactual responses}

    \BlankLine
    Initialize $\mathcal{B}_q\leftarrow\emptyset$ for every
    $q\in\mathcal{Q}_{\mathcal{E}}$\;

    \ForEach{$(x,m)\in\mathcal{D}_C$}{
        $z\leftarrow\mathcal{S}_{\mathcal{E}}(x)$\;
        \eIf{$z$ contains a valid explanation}{
            \ForEach{$q\in\mathcal{Q}_{\mathcal{E}}$}{
                $d\leftarrow g_{\mathcal{E},q}(x,z)$\;
                $\mathcal{B}_q\leftarrow
                \mathcal{B}_q\cup\{(d,m)\}$\;
            }
        }{
            Record an observable explanation failure with membership label
            $m$\;
        }
    }

    \BlankLine
    \ForEach{$q\in\mathcal{Q}_{\mathcal{E}}$}{
        $(\tau_q,\mathit{dir}_q)\leftarrow
        \mathrm{DTC}(\mathcal{B}_q)$\;
    }
    Select $q^{*}$ using calibration data only\;
    $\tau\leftarrow\tau_{q^{*}}$\;
    $\mathit{dir}^{*}\leftarrow\mathit{dir}_{q^{*}}$\;

    \BlankLine
    Initialize $\mathcal{R}\leftarrow\emptyset$\;
    \ForEach{$x\in\mathcal{D}_I$}{
        $z\leftarrow\mathcal{S}_{\mathcal{E}}(x)$\;
        \eIf{$z$ contains a valid explanation}{
            $d\leftarrow g_{\mathcal{E},q^{*}}(x,z)$\;
            $c_1\leftarrow
            (\mathit{dir}^{*}=\mathtt{leq})\land(d\leq\tau)$\;
            $c_2\leftarrow
            (\mathit{dir}^{*}=\mathtt{gt})\land(d>\tau)$\;
            \eIf{$c_1\lor c_2$}{
                $\widehat{m}\leftarrow 1$\;
            }{
                $\widehat{m}\leftarrow 0$\;
            }
        }{
            $\widehat{m}\leftarrow
            \mathrm{HandleFailure}(\mathcal{E})$\;
        }
        $\mathcal{R}\leftarrow
        \mathcal{R}\cup\{(x,\widehat{m})\}$\;
    }

    \BlankLine
    \Return{$\mathcal{R}$}\;
\end{algorithm}

%% file: sections/6-results.tex
\section{Results}
\label{sec:results}

In this section, we present our experimental results. Section~\ref{sec:experimental-setup} describes the experimental setup. Section~\ref{sec:lime-results} reports the results using LIME explanations and the corresponding calibration and distance analysis. Section~\ref{sec::tabular_cf} presents the counterfactual explanation results across three generators. Section~\ref{sec:extension-to-images} reports the extension to visual counterfactual explanations. Finally, Section~\ref{sec:comparison} compares \sys with prior work.


\subsection{Experimental Setup}
\label{sec:experimental-setup}

In our experiments, we evaluate \sys using both LIME~\cite{lime} and counterfactual explanations. We use the LIME tabular explainer. We additionally evaluate three counterfactual generators: NICE ~\cite{nice-counterfactual}, Wachter~\cite{counterfactual-watcher}, Growing Spheres~\cite{GS-counterfactual}. For a returned counterfactual $x'$, the attack uses its Euclidean displacement $\lVert x-x' \rVert$ from the queried sample $x$ as its membership signal. For a counterfactual request, the response contains the predicted label together with either a valid counterfactual $x'$ or an explicit generation failure. Operations required to construct an explanation are performed internally by the MLaaS service and therefore do not increase the attacker's external query count. Consequently, after calibration, \sys requires only one explanation request per target sample, regardless of the explanation mechanism used by the service. More details on the setup of the explainers can be found in Appendix~\ref{app:implementation-details}. Details about dataset splits can be found in Table~\ref{tab:dataset_splits} in Appnedix~\ref{app:dataset-splits}.

We evaluate \sys on four benchmark datasets introduced by Shokri et al.~\cite{shokri-mia} and widely adopted in subsequent literature: Purchase-100,\footnote{\url{https://github.com/privacytrustlab/datasets}\label{fn:privacy}} Location,\footref{fn:privacy} Texas-100,\footref{fn:privacy} and Adult.~\footnote{\url{http://archive.ics.uci.edu/ml/datasets/Adult}} For consistency, we instantiate the target model as a multi-layer perceptron (MLP) and keep the training pipeline fixed except for dataset-dependent choices. Each target MLP uses fully connected layers with 128 neurons and the \texttt{tanh} activation function. We vary the number of hidden layers from 1 to 4 depending on the dataset complexity. We train the models using the Adam optimizer and categorical cross-entropy loss, together with $L_2$ regularization and learning-rate decay. This setup has been used in Choquette-Choo et al.~\cite{choo-label-only} following the work of Shokri et al.~\cite{shokri-mia}.
Table~\ref{tab:dataset-statistics} summarizes the dataset statistics.

For the extension of \sys to images in \Cref{sec:extension-to-images}, we use CIFAR-10~\footnote{\url{https://www.cs.toronto.edu/~kriz/cifar.html}\label{fn:cifar}} and CIFAR-100.\footref{fn:cifar} For CIFAR-10, the target model is a custom three-block convolutional neural network (CNN) with two $3\times3$ convolutions per block. For CIFAR-100, we use a RestNet-18 with four residual stages of widths 64, 128, 256, and 512. DiVE uses a BigGAN-inspired $\beta$-total correlation variational autoencoder ($\beta$-TCVAE) as a generative model for the explainer with a 128-dimensional latent space: width/depth of $4/2$ for CIFAR-10 and $6/3$ for CIFAR-100. 

We evaluate the attack using standard metrics: Area Under the Receiver Operating Characteristic curve (ROC--AUC) and attack success rate (ASR) computed as $ASR = \frac{TP + TN}{TP + TN + FP + FN}$, where $TP, FP, TN, FN$ denote true/false positives and negatives.

\begin{table}[t]
\centering
\caption{Statistics of the datasets used in the LIME attack before preprocessing.}
\begin{tabular}{lcc}
\toprule
\textbf{Dataset} & \textbf{Dataset Size} & \textbf{\# Features} \\
\midrule
Purchase-100 & 197,324 & 600 \\
Texas-100    & 67,330  & 6,170 \\
Location     & 5,010   & 446 \\
Adult        & 48,842  & 14 \\
\bottomrule
\end{tabular}
\label{tab:dataset-statistics}
\end{table}



\subsection{LIME Explanations}
\label{sec:lime-results}
\label{sec:ablation}

We evaluate \sys across multiple datasets and target-model configurations. Figure~\ref{fig:attack-accuracy-best} reports ASRs of $0.85$ on Purchase-100, $0.71$ on Location, and $0.64$ on Texas-100. \sys performs best on Purchase-100 and Location, surpasses the baseline accuracy on Texas-100, and requires only one query after calibration. These results are consistent with the dataset and model characteristics. By contrast, the Adult model leaks less membership information, plausibly because of its smaller feature space and lack of a train-test gap. Adult has only 14 features, compared with 600, 446, and 6,170 for Purchase-100, Location, and Texas-100, respectively. In our experiments, this limited feature space improved generalization and reduced ASR, consistent with prior work~\cite{shokri-xai,salem2018-model_data_independent,choo-label-only}. We therefore did not pursue further ablations on Adult. The remainder of this section examines how calibration-set size affects \sys and how sample distances to the LIME-estimated decision boundary are distributed across datasets.

\input{graphs/best_accuracies}

\par\noindent\textbf{Calibration dataset ($\mathcal{D}_C$) size.}
\label{sec:ablation_calibration}
To evaluate the optimal calibration-dataset ($\mathcal{D}_C$) size, we conduct ablations on Purchase-100 and Location, where our method performs best. We use a fixed inference dataset ($\mathcal{D}_I$) and vary the size of $\mathcal{D}_C$ to measure its effect on attack performance.

\begin{figure}[htbp]
    \centering
    \begin{subfigure}[b]{0.48\columnwidth}
        \centering
        \includegraphics[width=\linewidth]{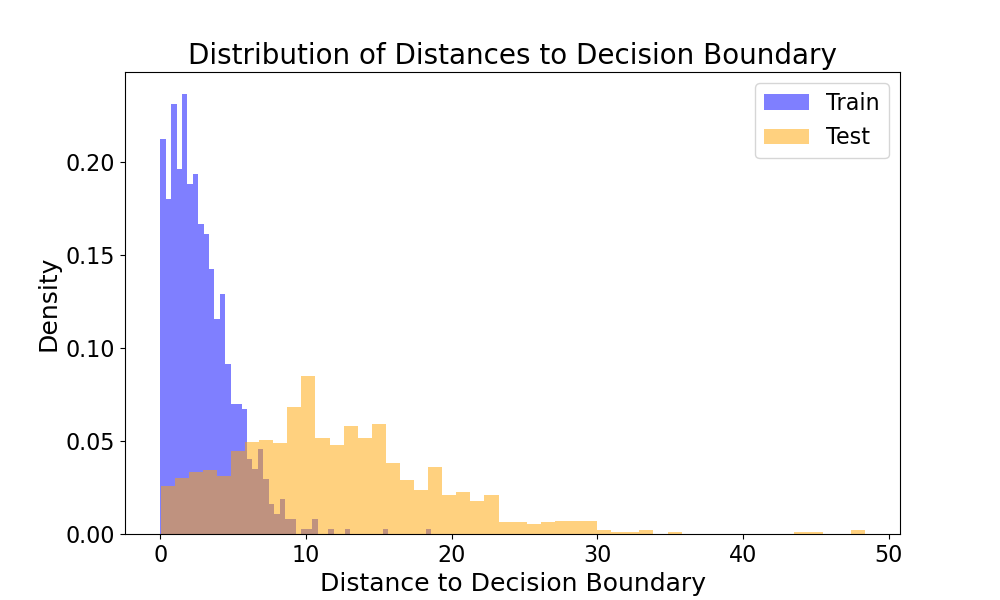}
        \caption{Less generalized model.}
        \label{fig:dists:overfitted}
    \end{subfigure}
    \hfill
    \begin{subfigure}[b]{0.48\columnwidth}
        \centering
        \includegraphics[width=\linewidth]{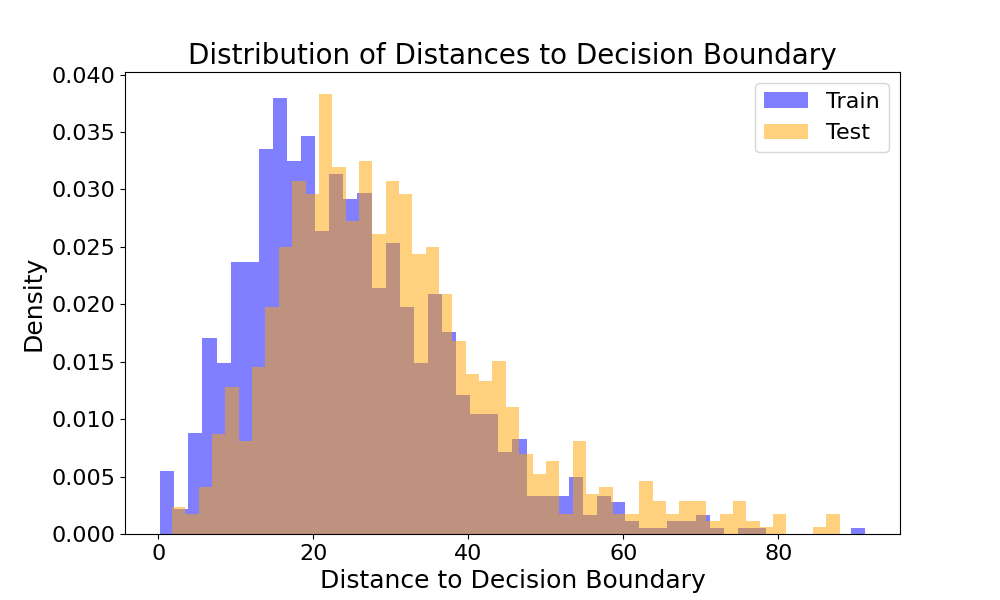}
        \caption{Well generalized model.}
        \label{fig:dists:generalized}
    \end{subfigure}
    \caption{Distance distribution of instances to the LIME-fitted local model based on the model generalization on Purchase-100 dataset.}
    \label{fig:distances}
\end{figure}

Figures~\ref{fig:sub_asr_calib_purchase} and~\ref{fig:sub_asr_calib_location} show the results for Purchase-100 and Location, respectively. In Figure~\ref{fig:sub_asr_calib_purchase}, calibration sets of 20 or fewer samples cause member and non-member ASRs to diverge markedly. With insufficient samples, threshold selection becomes unstable, placing the estimated distance threshold ($\tau$) either too close to zero or too far into the distribution's tail (see Figure~\ref{fig:dists:overfitted}).

With only 2 samples in $\mathcal{D}_C$, for example, the algorithm initially sets $\tau=10.76$. This yields high member ASRs but poor non-member ASRs. The inferred rule assumes that members have smaller distances than non-members, highlighting the need to estimate both the threshold and inequality direction accurately. As $\mathcal{D}_C$ grows, the threshold decreases to $\tau=10.33$, $\tau=6.72$, and $\tau=5.72$ for 10, 20, and 30 samples, respectively. At approximately 30 samples, both the threshold and direction stabilize, producing reliable ASR curves.

\input{graphs/asr_vs_calibration_size-combined}

\par
\begin{figure}[htbp]
    \centering
    \begin{subfigure}[b]{0.49\columnwidth}
        \centering
        \includegraphics[width=\linewidth]{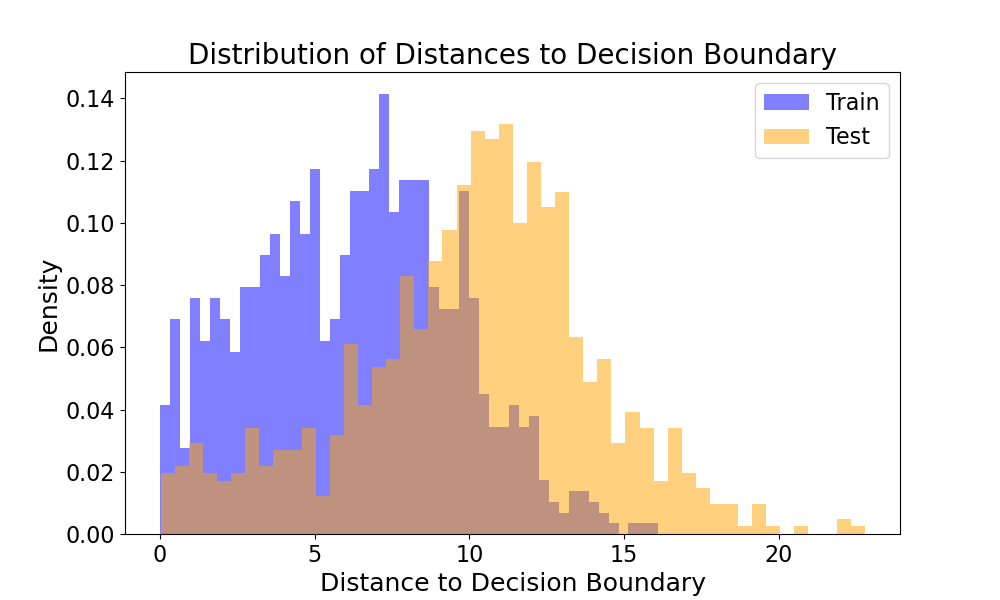}
        \caption{Location dataset}
        \label{fig:dists:location}
    \end{subfigure}
    \hfill
    \begin{subfigure}[b]{0.49\columnwidth}
        \centering
        \includegraphics[width=\linewidth]{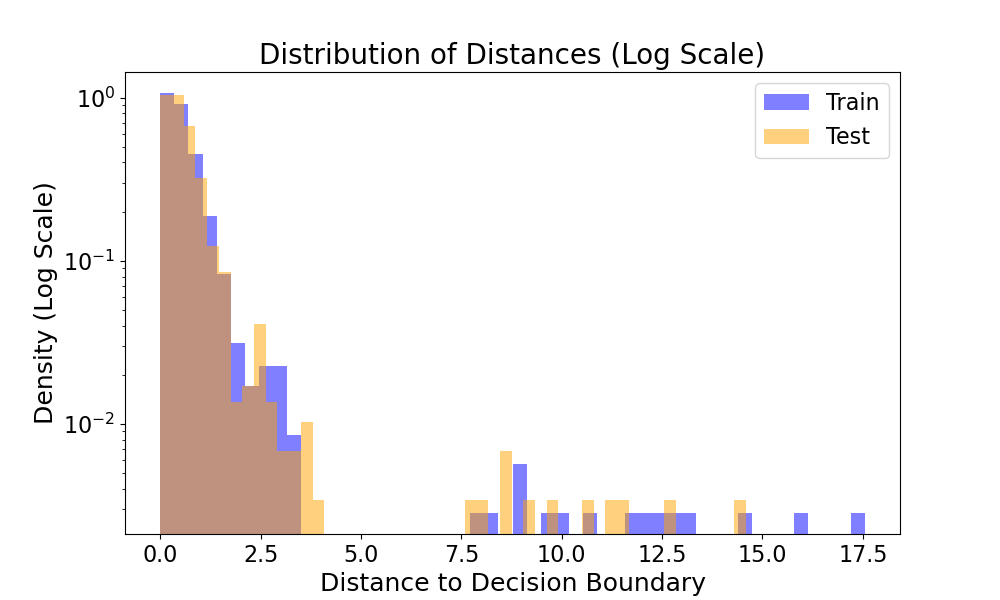}
        \caption{Adult dataset}
        \label{fig:dists:adult}
    \end{subfigure}
    \caption{Distance distribution of instances to the LIME-fitted local model on the (a) Location dataset and (b) Adult dataset.}
    \label{fig:distances2}
\end{figure}

\noindent\textbf{Distance to the boundary.} We also analyze the distance distributions from data samples to the local surrogate boundary estimated by LIME via \sys. Figures~\ref{fig:distances} and~\ref{fig:distances2} visualize these distances for the Purchase dataset, illustrating how the distance metric distinguishes members from non-members. Ultimately, ASR is driven by the separation between these distributions: greater separation minimizes classification errors, while substantial overlap makes threshold-based inference unreliable. 


Conversely, Figure~\ref{fig:dists:generalized} illustrates a more generalized target model where member and non-member distributions are nearly indistinguishable. Due to this severe overlap, no effective distance threshold can be established to separate the two groups.

Similar patterns across datasets support the same interpretation. Our target model trained on the Location dataset would create two different distance distributions (see Figure~\ref{fig:dists:location}). For the Adult dataset, however, the distribution for members and non-members shown in Figure~\ref{fig:dists:adult} is not practically separable, as the model generalizes well. This is consistent with the lowered ASR observed on Adult both in our experiments and in prior MIA studies~\cite{shokri-xai,salem2018-model_data_independent,choo-label-only}.

\subsection{Counterfactual Explanations}
\label{sec::tabular_cf}

In this section, we evaluate whether the calibrated $(\tau,\mathrm{dir})$ decision rule that \sys learns for LIME transfers, without modification, to a structurally different family of explanations. Counterfactual explanations provide a direct model-space displacement: a valid counterfactual $x'$ changes the target model's prediction, allowing us to measure the generator-selected displacement $\lVert x-x'\rVert$ without estimating a local surrogate boundary. 
\begin{figure}[t]
    \centering
    \includegraphics[width=\columnwidth]{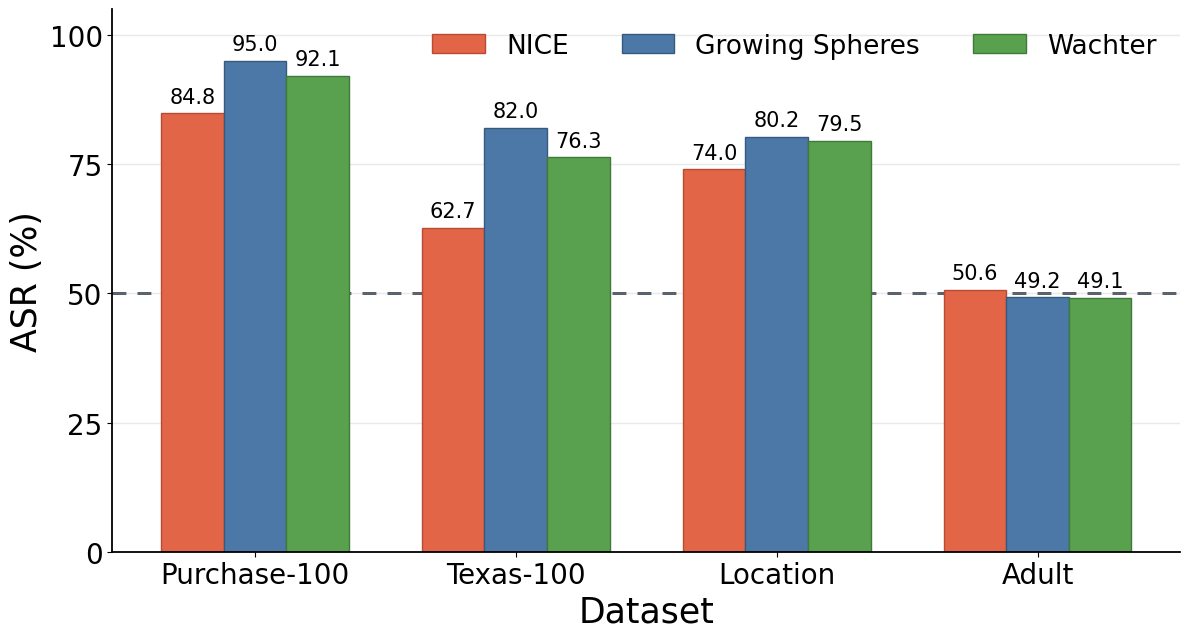}
    \caption{ASR across datasets for XposedMIA using the NICE, Growing Spheres, and Wachter counterfactuals. The dashed line denotes random guessing.}
    \label{fig:cf-asr}
\end{figure}
\input{graphs/asr_vs_calibration_size_cf}

\par\noindent\textbf{NICE.} With NICE explanations, our attack achieves balanced ASRs of 84.8\%, 74.0\%, and 62.7\% on Purchase-100, Location, and Texas-100, respectively, as reported in \Cref{fig:cf-asr}. Figure~\ref{fig:cf-nice-location-dist} in Appendix~\ref{app:distances} shows the member and non-member counterfactual-distance distributions on Location, with a calibrated threshold of $\tau=8.18$. NICE achieves complete counterfactual coverage on all evaluated datasets, so generation failures do not contribute to its membership signal. However, the weaker performance on Texas-100 shows that complete coverage is insufficient when counterfactual distances do not strongly separate members from non-members. Because NICE uses nearest unlike reference instances, this effect may be amplified in the high-dimensional Texas-100 feature space, where reference-set proximity may not closely track distance to the model boundary.

\par\noindent\textbf{Wachter.} With Wachter counterfactuals, our attack achieves balanced ASRs of 92.1\%, 79.5\%, and 76.3\% on Purchase-100, Location, and Texas-100, respectively, as reported in \Cref{fig:cf-asr}. Figure~\ref{fig:cf-wachter-purchase-dist} in Appendix~\ref{app:distances} shows the corresponding member and non-member counterfactual-distance distributions on Purchase-100, with $\tau=1.94$. On Adult, we use a stronger penalty schedule to increase counterfactual coverage, achieving $100.0\%$ member/non-member coverage but only $49.4\%$ ROC-AUC. Thus, reliable counterfactual generation is necessary but insufficient for effective membership separation.
This generator-side configuration does not change the attacker's access model: the attacker still receives one counterfactual response and applies the same calibrated $(\tau,\mathrm{dir})$ decision rule.

\par\noindent\textbf{Growing Spheres.}
The attack on Growing Spheres achieves balanced ASRs of 95.0\%, 82.0\%, and 80.16\% on Purchase-100, Texas-100, and Location, respectively. Figure~\ref{fig:cf-asr} and~\ref{fig:cf-gs-texas-dist}, Appendix~\ref{app:distances} illustrate the ASR and member and non-member counterfactual-distance distributions on Texas-100, for which $\tau=23.65$. For Texas-100, we therefore increase the search radius to recover valid counterfactuals over a broader region of the feature space.

\par\noindent\textbf{Cross-generator analysis.} Table~\ref{tab:cf-results} shows that counterfactual coverage alone does not determine attack effectiveness. NICE achieves $100\%$ coverage on every dataset, yet Growing Spheres often yields stronger membership separation despite lower coverage. Thus, MIA effectiveness depends not only on valid counterfactual generation but also on how strongly the induced distance distributions separate members from non-members. Because the generators construct counterfactuals differently, they expose different aspects of the decision geometry: NICE is shaped by its reference data, Wachter by its optimization objective, and Growing Spheres by its radial search.

The Adult results reinforce this point: despite complete coverage in most settings, all three generators remain close to random guessing when member and non-member distances are insufficiently separated. Conversely, in most incomplete-coverage settings, valid counterfactuals are returned less often for members than non-members, consistent with members lying farther from the decision boundary and requiring larger displacements within a finite search budget. Generation failure can therefore provide an additional membership signal alongside the distance of successfully returned counterfactuals.

\begin{table}[t]
    \centering
    \caption{Counterfactual-distance results under the calibrated DTC rule. CF success reports valid label-changing counterfactuals for members (M) and non-members (N); each attack uses one generator request per target.}
    \label{tab:cf-results}
    \small
    \setlength{\tabcolsep}{4pt}

    \resizebox{\columnwidth}{!}{%
    \begin{tabular}{llrrrr}
        \toprule
        Dataset & Generator & ASR & ROC--AUC & \multicolumn{2}{c}{CF Success} \\
        \cmidrule(lr){5-6}
         & & & & M & N \\
        \midrule
        \multirow{3}{*}{Purchase-100} & NICE & 84.8\% & 91.6\% & 100.0\% & 100.0\% \\
         & Growing Spheres & 95.0\% & 97.6\% & 88.7\% & 98.9\% \\
         & Wachter & 92.1\% & 96.7\% & 99.6\% & 100.0\% \\
        \midrule
        \multirow{3}{*}{Texas-100} & NICE & 62.7\% & 67.6\% & 100.0\% & 100.0\% \\
         & Growing Spheres & 82.0\% & 88.1\% & 78.4\% & 95.9\% \\
         & Wachter & 76.3\% & 79.8\% & 85.9\% & 98.3\% \\
        \midrule
        \multirow{3}{*}{Location} & NICE & 74.0\% & 81.2\% & 100.0\% & 100.0\% \\
         & Growing Spheres & 80.2\% & 87.1\% & 81.6\% & 93.2\% \\
         & Wachter & 79.5\% & 84.7\% & 99.7\% & 99.8\% \\
        \midrule
        \multirow{3}{*}{Adult} & NICE & 50.6\% & 52.1\% & 100.0\% & 100.0\% \\
         & Growing Spheres & 49.2\% & 49.1\% & 98.9\% & 99.1\% \\
         & Wachter & 49.1\% & 49.4\% & 100.0\% & 100.0\% \\
        \bottomrule
    \end{tabular}%
    }
\end{table}
\begin{figure}[t]
    \centering
    \includegraphics[width=0.9\columnwidth]{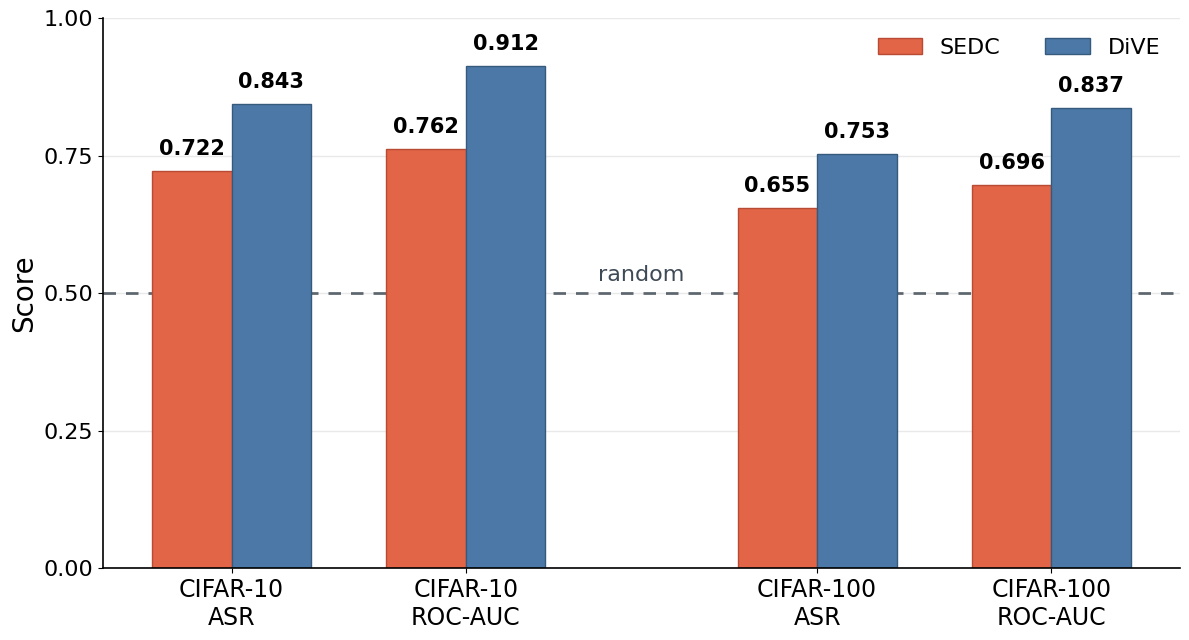}
    \caption{Results using the SEDC and DiVE counterfactual explanations on CIFAR-10 and CIFAR-100. Both ASR and ROC--AUC are reported. The results present substantial membership information leakage above the random baseline.}
    \label{fig:sedc-dive-results}
\end{figure}

\par\noindent\textbf{Calibration-size ablation.} Figure~\ref{fig:cf-calibration-ablation} repeats the calibration analysis of Section~\ref{sec:lime-results} for counterfactual distances, resampling only $\mathcal{D}_C$ while keeping the target model, inference records, and generator responses fixed; each point averages 30 nested resamples. As in LIME, very small calibration sets produce unstable estimates of both the threshold $\tau$ and the inequality direction, which appears as a pronounced imbalance between member and non-member ASR. For example, with only two calibration records per membership class, Wachter on Purchase-100 achieves mean member and non-member ASRs of $98.8\%$ and $73.6.\%$, respectively, with $\tau=1.18\pm0.52$. Increasing the calibration size to 10, 20, and 30 records from each membership class yields thresholds of $\tau=1.72\pm0.37$, $\tau=1.88\pm0.18$, and $\tau=1.90\pm0.16$, respectively, while the corresponding member/non-member ASRs become $94.5\%/89.3\%$, $94.1\%/91.9\%$, and $93.5\%/92.1.\%$. At 30 records per membership class, the balanced ASR reaches $92.8\%$, already close to the full-calibration value of $92.1\%$, with the threshold approaching its full-calibration value of $\tau=1.94$. The same stabilization pattern is observed for NICE and Growing Spheres. Across all generators, roughly 20--30 calibration records per class are sufficient for the DTC rule to reach a stable operating point, after which additional data produces only marginal changes. Together with the LIME results in Section~\ref{sec:lime-results}, this suggests that the small calibration requirement is primarily a property of the DTC procedure rather than of the underlying explanation mechanism. We also perform an ablation to study the effect of changing the second stage of DTC with an objective that maximizes balanced accuracy instead of minimizing the difference between member and nonmember accuracy. Our results are summarized in Table~\ref{tab:threshold_calibration_ablation}, Appendix~\ref{app:dtc-ablation}.

\subsection{Extension to Images}
\label{sec:extension-to-images}
To evaluate whether \sys generalizes to more complex datasets, we extend our experiments to image classifiers. We use CIFAR-10 and CIFAR-100 since they are among the main benchmark datasets for most MIA works~\cite{choo-label-only, carlini_membership_2022, nasr_whitebox_mia, salem2018-model_data_independent, li_enhanced_2025}. We target two visual counterfactual explanation methods: SEDC~\cite{vermeire-sedc-image} and DiVE~\cite{dive-rodriguez}. We select these two methods to introduce diversity in counterfactual generation and demonstrate the generalizability of our approach. SEDC changes the prediction by removing image segments, while DiVE utilizes a generative model (VAE) and searches in its latent space for counterfactuals. Example SEDC and DiVE counterfactuals can be seen in Appendix~\ref{app:visual-cf-exampe}. It is important to note that the attack itself remains unchanged once the distance scores are obtained; we use the same DTC and MIA framework described in \Cref{sec:method}. 

\par\noindent\textbf{Relevance to privacy-sensitive applications.}
CIFAR provides a controlled benchmark for testing whether visual
counterfactuals expose membership-related signals. The same interface
can arise in privacy-sensitive image-classification services, such as
medical imaging.

\par\noindent\textbf{SEDC.}
\Cref{fig:sedc-dive-results} demonstrates the balanced ASR using SEDC counterfactuals and the ROC--AUC results for the CIFAR-10 and CIFAR-100 datasets. \sys using SEDC reports 0.722 and 0.655 ASR for CIFAR-10 and CIFAR-100, respectively. The reported results show membership information leakage.

\begin{figure}[t]
    \centering
    \includegraphics[width=\columnwidth]{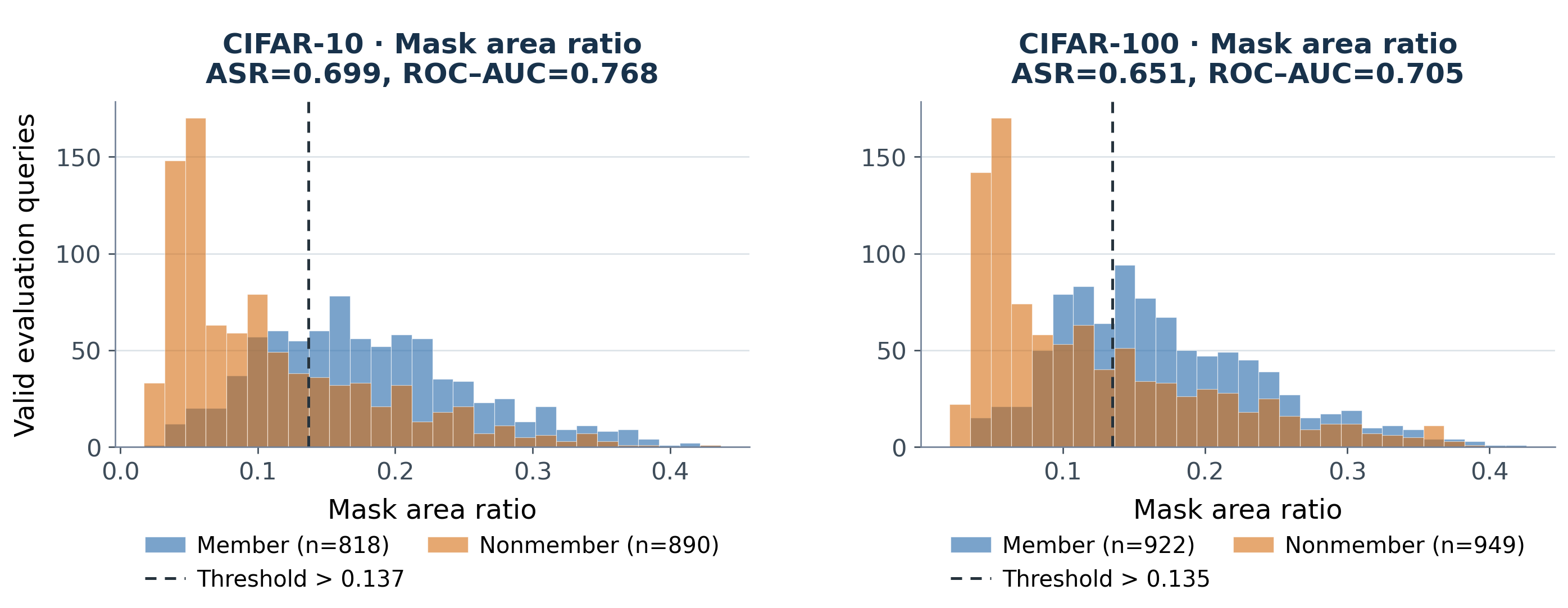}
    \caption{CIFAR-10 and CIFAR-100 mask area ratio distances and the chosen decision threshold for the SEDC attack.}
    \label{fig:sedc-distances}
\end{figure}

\Cref{fig:sedc-distances} visualizes the distance distributions for members and non-members. The DTC-derived threshold from \Cref{sec:calibration} is marked by a vertical line. The members are skewed to the right of the plot since they have higher distances, while non-members are skewed to the left. The members that are on the left of the line are the false negatives, while the non-members that are on the right of the threshold line are the false positives.

\par\noindent\textbf{DiVE.}
DiVE uses a generative model to generate counterfactuals. Since DiVE is targeted by nature, it finds a set of counterfactuals in the latent space of the generative model that would be classified as the target class. We then pick the counterfactual that has the smallest proximity score $s_j=\lambda_{\mathrm{prox}}R_j+\lambda_{\mathrm{latent}}L_j$, where $R_j$ is the sum of the per-channel mean absolute pixel differences between the factual image and candidate $j$, and $L_j=\lVert\epsilon_j\rVert_1$ measures the magnitude of its latent-space perturbation. 

\Cref{fig:sedc-dive-results} demonstrates the results for CIFAR-10 and CIFAR-100. We observe the best results using SSIM distance, with a balanced accuracy of $0.843$ and ROC--AUC of $0.912$ for CIFAR-10 and a balanced accuracy of $0.753$ and ROC--AUC of $0.837$ for CIFAR-100.

\begin{figure}[t]
    \centering
    \includegraphics[width=\columnwidth]{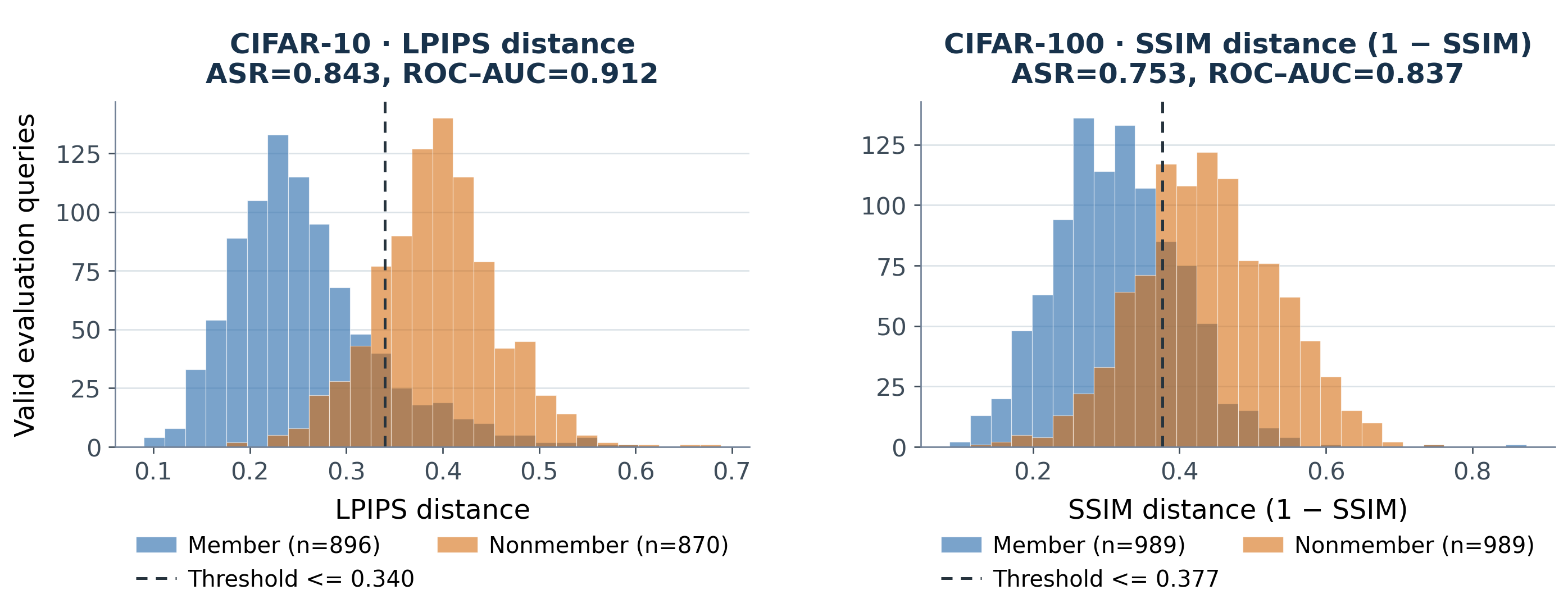}
    \caption{CIFAR-10 Learned Perceptual Image Patch Similarity (LPIPS) and CIFAR-100 SSIM distances and the chosen decision threshold for the DiVE attack.}
    \label{fig:dive-distances}
\end{figure}

\Cref{fig:dive-distances} presents clear separation between members and non-members in terms of their distance to their counterfactuals. Yet, there are some false positives and false negatives, as detailed for SEDC. Since the ASR is higher for DiVE explanations, we observe a cleaner and more separated distribution compared to the SEDC.

\subsection{Comparison to Other Works} 
\label{sec:comparison}

We compare \sys with seven representative MIA approaches: (i) Gap Attack~\cite{yeom-overfitting}, (ii) Choquette-Choo et al.~\cite{choo-label-only}, (iii) Shokri et al.~\cite{shokri-xai}, (iv) Ye et al.~\cite{ye_classical_mia}, (v) YOQO~\cite{yoqo-mia}, (vi) DHAttack~\cite{li_enhanced_2025}, and (vii) LoMime~\cite{lomime}. We report the ASR and ROC--AUC across the Purchase-100, Location, and Texas-100 datasets in Table~\ref{tab:comparison}, Appendix~\ref{app:comparison}. Compared with the Gap Attack~\cite{yeom-overfitting} and prior explanation-variance attacks~\cite{shokri-xai}, \sys achieves stronger results while using only a single query after calibration and requiring no shadow-model training. Unlike shadow-model-based methods, \sys uses its auxiliary data only for threshold calibration rather than for training surrogate models, which substantially reduces the amount of data and computational cost for the attack. Furthermore, compared with perturbation-based boundary-distance attacks such as~\cite{choo-label-only}, \sys substantially reduces the attack-time query budget while still achieving strong and comparable attack performance.

\paragraph{Performance under constrained query budgets.} Table~\ref{tab:comparison} reports the query budgets associated with prior attacks' reported performance. Choquette-Choo et al.~\cite{choo-label-only} require approximately $2{,}500$ queries per candidate for their boundary-distance attack to match their strongest confidence-based baseline, and approximately $12{,}500$ for their combined attack. With $<100$ queries, the improvement over the gap attack is only approximately four percentage points, while below $300$ queries the simpler noise-robustness attack performs better because the boundary estimate remains imprecise. LoMime~\cite{lomime} likewise degrades under a reduced one-time model-extraction budget: on Purchase-100, AUC increases from approximately $0.57$ at $10^4$ queries to $0.87$ at $10^6$, and on Texas-100 from approximately $0.52$ to $0.81$ over the same budgets. Thus, the performance of query-intensive attacks can depend substantially on the available budget. This does not apply uniformly to all prior work: DHAttack~\cite{li_enhanced_2025} and YOQO~\cite{yoqo-mia} reduce target-model queries but rely on additional reference/shadow models or auxiliary computation.


\paragraph{\textbf{Future Directions.}} As future work, we will focus on a more comprehensive set of explanation tools, deeper neural network architectures, and extend our analysis to text data, their corresponding explanations, and LLMs. A more detailed analysis of image XAI leakage across datasets and explainers is another promising direction for future research. Finally we plan on building a unified framework that combines attacks for tabular, image, and text data.

%% file: graphs/best_accuracies.tex
\begin{figure}[t]
    \definecolor{predcolor}{RGB}{102, 51, 153}   
    \definecolor{basecolor}{RGB}{220, 100, 70}   
    \scalebox{0.95}{
    \begin{tikzpicture}
    \begin{axis}[
        ybar,
        bar width=9pt,
        width=7.5cm,
        height=4.5cm,
        ymin=0, ymax=1.0,
        ytick={0, 0.2, 0.4, 0.5, 0.6, 0.7, 0.8, 0.9, 1.0},
        yticklabels={0\%, 20\%, 40\%, 50\%, 60\%, 70\%, 80\%, 90\%, 100\%},
        ylabel={\large Attack Accuracy},
        ylabel style={font=\large},
        xtick=data,
        xticklabels={Texas, Purchase, Location, Adult},
        xticklabel style={font=\large},
        tick label style={font=\normalsize},
        ymajorgrids=true,
        grid style={dashed, gray!40},
        legend style={
            at={(0.5,-0.22)},
            anchor=north,
            legend columns=2,
            font=\normalsize,
            column sep=1.5em,
            draw=none,
        },
        legend cell align=left,
        extra y ticks={0.5},
        extra y tick style={
            grid=major,
            grid style={dashed, black, thick},
            tick label style={font=\normalsize},
        },
        extra y tick labels={},
        axis line style={-},
        clip=false,
    ]
    
    
    \addplot[fill=predcolor, draw=none] coordinates {
        (1, 0.6171)   
        (2, 0.8657)   
        (3, 0.6857)   
        (4, 0.505)   
    };
    \addlegendentry{Members}

    \addplot[fill=basecolor, draw=none] coordinates {
        (1, 0.67)   
        (2, 0.84)   
        (3, 0.7397) 
        (4, 0.51) 
    };
    \addlegendentry{Non-Members}

    \node[anchor=south east, font=\footnotesize] at (rel axis cs:0.98, 0.5) {Random guess};
    
    \end{axis}
    \end{tikzpicture}}
    \vspace{-0.5em}
    \caption{Attack accuracy for members and non-members for Location, Purchase-100, Texas-100, and Adult datasets.}
    \label{fig:attack-accuracy-best}
\end{figure}

%% file: graphs/asr_vs_calibration_size-combined.tex
\begin{figure}[!htbp]
    \centering
    
    \begin{subfigure}{0.45\textwidth}
        \centering
        \begin{tikzpicture}
            \begin{axis}[
                width=\linewidth, 
                height=3.7cm,     
                xlabel={Calibration Size},
                ylabel={ASR},
                label style={font=\small}, 
                tick label style={font=\footnotesize}, 
                legend pos=south east,
                legend style={font=\scriptsize, nodes={scale=0.9, transform shape}}, 
                grid=major,
                ytick={0.5, 0.6, 0.7, 0.8, 0.9, 1.0}, 
                xtick={30, 200, 500, 1000},
                ymin=0.4, ymax=1.0,
                xmin=0, xmax=1200,
            ]
            
            \addplot[
                color=blue,
                mark=*,
                thick,
                mark size=1.5pt 
            ] coordinates {
                (2, 0.99571)
                (10, 0.99571)
                (20, 0.96)
                (30, 0.90714)
                (40, 0.90857)
                (50, 0.89714)
                (100, 0.87)
                (200, 0.85714)
                (300, 0.85714)
                (400, 0.81)
                (500, 0.82286)
                (600, 0.82286)
                (700, 0.82286)
                (800, 0.84857)
                (900, 0.85714)
                (1000, 0.87)
                (1100, 0.87429)
                (1200, 0.87429)
            };
            \addlegendentry{Members}
            
            \addplot[
                color=red,
                mark=square*,
                thick,
                mark size=1.5pt
            ] coordinates {
                (2, 0.54857)
                (10, 0.57286)
                (20, 0.76286)
                (30, 0.80714)
                (40, 0.80429)
                (50, 0.81429)
                (100, 0.83714)
                (200, 0.84571)
                (300, 0.84571)
                (400, 0.86857)
                (500, 0.85714)
                (600, 0.85714)
                (700, 0.86143)
                (800, 0.84714)
                (900, 0.84571)
                (1000, 0.83714)
                (1100, 0.83143)
                (1200, 0.83)
            };
            \addlegendentry{Non-Members}
            
            \end{axis}
        \end{tikzpicture}
        \caption{Purchase-100 dataset}
        \label{fig:sub_asr_calib_purchase}
    \end{subfigure}
    \begin{subfigure}{0.45\textwidth}
        \centering
        \begin{tikzpicture}
            \begin{axis}[
                width=\linewidth, 
                height=3.7cm,
                xlabel={Calibration Size},
                ylabel={ASR},
                label style={font=\small},
                tick label style={font=\footnotesize},
                legend pos=south east,
                legend style={font=\scriptsize, nodes={scale=0.9, transform shape}},
                grid=major,
                ytick={0.4,0.5, 0.6, 0.7, 0.8, 0.9, 1.0},
                xtick={30, 200, 500, 1000},
                ymin=0.3, ymax=0.9,
                xmin=0, xmax=1200,
            ]
            
            \addplot[
                color=blue,
                mark=*,
                thick,
                mark size=1.5pt
            ] coordinates {
                (10, 0.41587)
                (30, 0.61587)
                (60, 0.74286)
                (120, 0.78571)
                (180, 0.80317)
                (240, 0.79365)
                (300, 0.80317)
                (360, 0.80159)
                (420, 0.7873)
                (480, 0.79524)
                (540, 0.7873)
                (600, 0.76984)
                (660, 0.78571)
                (720, 0.79206)
                (780, 0.79206)
                (840, 0.79365)
                (900, 0.80159)
                (960, 0.80317)
                (1020, 0.80317)
                (1140, 0.80794)
            };
            \addlegendentry{Members}
            
            \addplot[
                color=red,
                mark=square*,
                thick,
                mark size=1.5pt
            ] coordinates {
                (10, 0.81111)
                (30, 0.65079)
                (60, 0.69206)
                (120, 0.67302)
                (180, 0.65397)
                (240, 0.66825)
                (300, 0.65397)
                (360, 0.65714)
                (420, 0.67302)
                (480, 0.66667)
                (540, 0.67302)
                (600, 0.67937)
                (660, 0.67302)
                (720, 0.66984)
                (780, 0.66984)
                (840, 0.66984)
                (900, 0.65556)
                (960, 0.65397)
                (1020, 0.65397)
                (1140, 0.64762)
            };
            \addlegendentry{Non-Members}
            
            \end{axis}
        \end{tikzpicture}
        \caption{Location dataset}
        \label{fig:sub_asr_calib_location}
    \end{subfigure}
    \vspace{-0.5em}
    \caption{Comparison of the attack accuracies on members and non-members for different calibration sizes on the (a) Purchase-100 and (b) Location datasets.}
    \label{fig:asr_vs_calibration_combined}
\end{figure}

%% file: graphs/asr_vs_calibration_size_cf.tex
%

\begin{figure*}[!t]
    \centering

    \pgfplotsset{
        cf calibration axis/.style={
            width=\linewidth,
            height=3.5cm,
            xlabel={Calibration Size},
            ylabel={ASR},
            xmin=0,
            xmax=310,
            xtick={0,100,200,300},
            grid=major,
            grid style={gray!35},
            axis line style={black},
            label style={font=\small},
            tick label style={font=\footnotesize},
            every axis plot/.append style={thick},
        },
    }

    \begin{subfigure}[t]{0.33\textwidth}
        \centering
        \begin{tikzpicture}
            \begin{axis}[
                cf calibration axis,
                ymin=0.6,
                ymax=1.0,
                ytick={0.6,0.7,0.8,0.9,1.0},
                legend to name=cf-calibration-legend,
                legend columns=2,
                legend style={
                    font=\footnotesize,
                    /tikz/every even column/.append style={column sep=8pt},
                },
            ]
                \addplot[
                    color=blue,
                    mark=*,
                    mark size=1.5pt,
                ] coordinates {
                    (2, 0.98833)
                    (10, 0.94524)
                    (20, 0.94124)
                    (30, 0.93486)
                    (40, 0.93271)
                    (50, 0.94076)
                    (60, 0.93995)
                    (100, 0.93229)
                    (120, 0.92705)
                    (180, 0.92648)
                    (200, 0.92462)
                    (240, 0.92257)
                    (270, 0.92186)
                    (300, 0.91857)
                };
                \addlegendentry{Members}

                \addplot[
                    color=red,
                    mark=square*,
                    mark size=1.5pt,
                ] coordinates {
                    (2, 0.73624)
                    (10, 0.89338)
                    (20, 0.91867)
                    (30, 0.92081)
                    (40, 0.92186)
                    (50, 0.91838)
                    (60, 0.91852)
                    (100, 0.91871)
                    (120, 0.91990)
                    (180, 0.91957)
                    (200, 0.92095)
                    (240, 0.92229)
                    (270, 0.92248)
                    (300, 0.92429)
                };
                \addlegendentry{Non-Members}
            \end{axis}
        \end{tikzpicture}
        \caption{Wachter (Purchase-100)}
        \label{fig:cf-calibration-wachter}
    \end{subfigure}
    \hfill
    \begin{subfigure}[t]{0.33\textwidth}
        \centering
        \begin{tikzpicture}
            \begin{axis}[
                cf calibration axis,
                ymin=0.5,
                ymax=0.9,
                ytick={0.5,0.6,0.7,0.8,0.9},
            ]
                \addplot[
                    color=blue,
                    mark=*,
                    mark size=1.5pt,
                ] coordinates {
                    (2, 0.79767)
                    (10, 0.72159)
                    (20, 0.70783)
                    (30, 0.73000)
                    (40, 0.72508)
                    (50, 0.72534)
                    (60, 0.72487)
                    (100, 0.72095)
                    (120, 0.72016)
                    (180, 0.72439)
                    (200, 0.72751)
                    (240, 0.72418)
                    (270, 0.72540)
                };

                \addplot[
                    color=red,
                    mark=square*,
                    mark size=1.5pt,
                ] coordinates {
                    (2, 0.61212)
                    (10, 0.73767)
                    (20, 0.76233)
                    (30, 0.75037)
                    (40, 0.75550)
                    (50, 0.75455)
                    (60, 0.75397)
                    (100, 0.75519)
                    (120, 0.75608)
                    (180, 0.75439)
                    (200, 0.75291)
                    (240, 0.75413)
                    (270, 0.75397)
                };
            \end{axis}
        \end{tikzpicture}
        \caption{NICE (Location)}
        \label{fig:cf-calibration-nice}
    \end{subfigure}
    \hfill
    \begin{subfigure}[t]{0.33\textwidth}
        \centering
        \begin{tikzpicture}
            \begin{axis}[
                cf calibration axis,
                ymin=0.6,
                ymax=0.9,
                ytick={0.6,0.7,0.8,0.9},
            ]
                \addplot[
                    color=blue,
                    mark=*,
                    mark size=1.5pt,
                ] coordinates {
                    (2, 0.86505)
                    (10, 0.84576)
                    (20, 0.84676)
                    (30, 0.85729)
                    (40, 0.86057)
                    (50, 0.86019)
                    (60, 0.85919)
                    (100, 0.85210)
                    (120, 0.85171)
                    (180, 0.85505)
                    (200, 0.85681)
                    (240, 0.85705)
                    (270, 0.85943)
                    (300, 0.86000)
                };
    
                \addplot[
                    color=red,
                    mark=square*,
                    mark size=1.5pt,
                ] coordinates {
                    (2, 0.63710)
                    (10, 0.77290)
                    (20, 0.78429)
                    (30, 0.78186)
                    (40, 0.78195)
                    (50, 0.78295)
                    (60, 0.78333)
                    (100, 0.78643)
                    (120, 0.78600)
                    (180, 0.78381)
                    (200, 0.78286)
                    (240, 0.78248)
                    (270, 0.78133)
                    (300, 0.78000)
                };
            \end{axis}
        \end{tikzpicture}
        \caption{Growing Spheres (Texas-100)}
        \label{fig:cf-calibration-growing-spheres}
    \end{subfigure}

    \par\smallskip
    \pgfplotslegendfromname{cf-calibration-legend}
\vspace{-0.5em}
    \caption{Calibration-size sensitivity for three counterfactual
    generators. The panels report member and non-member attack success rates
    for representative Wachter, NICE, and Growing Spheres runs on
    Purchase-100, Location, and Texas-100, respectively. Each point is the mean
    over 30 nested calibration-subset resamples with a fixed inference set.}
    \label{fig:cf-calibration-ablation}
\end{figure*}

%% file: sections/8-conclusion.tex
\section{Conclusion}
\label{sec:conclusion}

In this paper, we presented \sys, a lightweight, \textbf{one-shot explanation-based black-box MIA framework}. We instantiate this framework with LIME and extend it to five counterfactual explanation methods spanning tabular and image domains. \sys either exploits the geometry of a local linear surrogate to approximate the decision boundary or uses the displacement between a factual sample and its returned label-changing counterfactual as a proxy for distance to the decision boundary.

Our evaluation across these structurally different explanation mechanisms demonstrates that the boundary-related information exposed through explanations can consistently reveal membership information. Moreover, the effectiveness of \sys across both tabular and image domains shows that this privacy leakage is not limited to a particular explanation method or data modality, making it relevant to many MLaaS platforms that release XAI explanations.
